\documentclass[11pt]{article}

\usepackage[preprint]{acl}

\usepackage{times}
\usepackage{latexsym}

\usepackage[T1]{fontenc}

\usepackage[utf8]{inputenc}

\usepackage{microtype}

\usepackage{inconsolata}

\usepackage{graphicx}
\usepackage{booktabs}
\usepackage{rotating}
\usepackage{amsmath} 
\usepackage{arydshln}
\usepackage{pifont}
\usepackage{multirow}
\newcommand{\cmark}{\ding{51}}
\newcommand{\xmark}{\ding{55}}

\usepackage{svg}
\title{Self-Stigma Is Not a Monolith, but Generic Empathy Is: Persona-Conditioned LLM Support for People Who Use Drugs}

\author{Layla Bouzoubaa and Rezvaneh Rezapour \\
  Department of Information Science \\
  Drexel University \\
  Philadelphia, PA \\
  \texttt{\{layla.bouzoubaa, shadi.rezapour\}@drexel.edu} \\
  }

\begin{document}
\maketitle
\begin{abstract}
Self-stigma predicts treatment avoidance and disengagement among people who use drugs (PWUD), yet conversational systems aiming to provide support typically treat self-stigma expression as a uniform signal. We present a three-phase, proof-of-concept study of a persona-aware approach to LLM support. Latent Profile Analysis (LPA) on indicator-level features from 1,174 self-stigma expressors on Reddit yields a four-persona typology validated against held-out behavioral and linguistic features. Sequential Bayesian and recurrent neural classifiers recover these personas from limited posting histories, substantially outperforming batch and few-shot LLM baselines (macro-F1 = 0.74 at 30 posts). Evaluation by eight clinical experts across three contemporary LLMs revealed a misalignment: persona-matched responses successfully achieved targeted behavioral shifts, yet raters holistically preferred the generic empathy of the persona-neutral baseline. Our findings suggest that holistic empathy judgments and clinically-aligned response design can pull in opposite directions, and that evaluating LLM-based stigma support requires rubrics capable of decomposing the two.


\end{abstract}

\section{Introduction}\label{sec:intro}

Self-stigma is the internalization of negative societal beliefs, often expressed as shame, self-blame, and a sense of being unworthy of help~\citep{corrigan2006self, livingstonCorrelatesConsequencesInternalized2010}. For people who use drugs (PWUD), self-stigma is a critical predictor of behavior change, directly influencing help-seeking and recovery pathways~\citep{corrigan2009whytry}. Those who might benefit most from support are often the most vulnerable to disengaging once they try.

These internal barriers leave linguistic traces on social media \citep{chenExaminingStigmaRelating2022, bouzoubaa_stigma_2024}. Because word choice reflects stable and situational psychological states~\citep{pennebaker2003psychological}, self-stigma expression in substance-use communities on Reddit can be reliably extracted and decomposed into theoretically grounded indicators~\citep{chenExaminingStigmaRelating2022, roeslerTheoryinformedDeepLearning2024, wangIdentifyingStigmaPhenotypes2025, bouzoubaa2026phenotypes}. This signal opens a path toward systems that detect self-stigma and generate responses calibrated to how a specific user experiences it.

This capability is highly relevant as telehealth expands as a delivery model for opioid use disorder (OUD) treatment~\citep{calesCOVID19PandemicOpioid2022, georgiadisTelemedicationOpioidUse2023}. Remote interventions risk compounding existing stigma if they assume a homogeneous user~\citep{couchPatientPerceptionsExperiences2024}. Conversational agents and large language models (LLMs) face a related risk \citep{mooreLLMsTherapist2025}. Without sensitivity to how self-stigma manifests, well-intentioned automated support can easily be perceived as dismissive or pathologizing. The goal is to deliver just-in-time support at the moment of self-devaluation, before downstream harms accumulate~\citep{corrigan2006self}, which requires recognizing the specific user profile first.

\begin{figure*}
    \centering
    \includegraphics[width=0.8\linewidth]{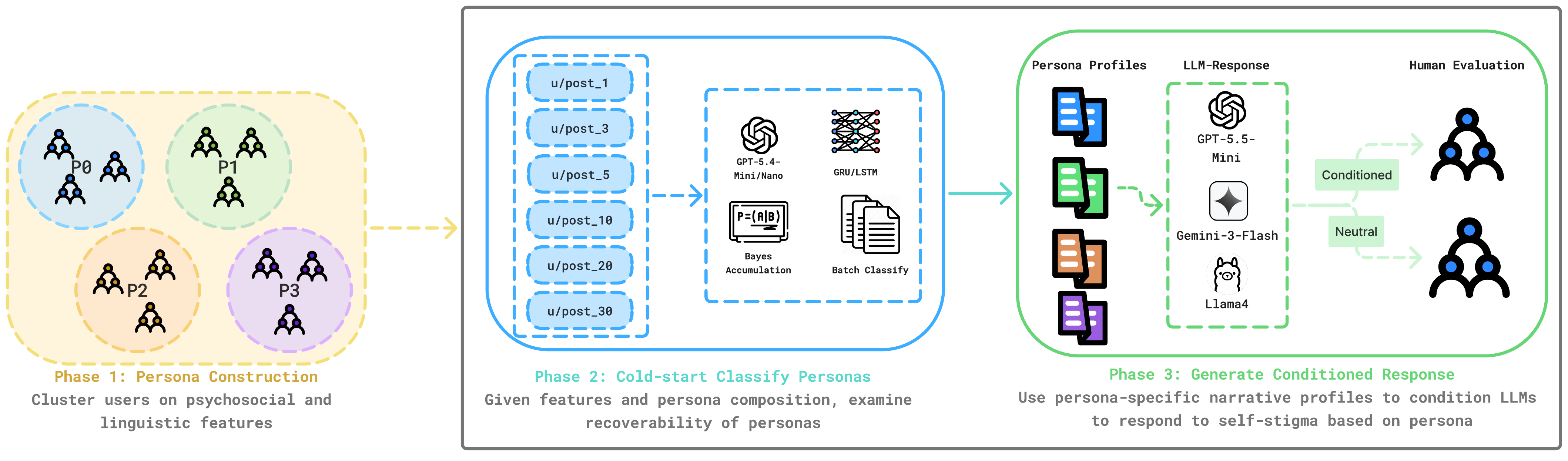}
    \caption{Overview of our three-phase approach. Phases~2--3 (outer box) constitute the proposed support approach; Phase~1 establishes the persona typology it operates over.}
    \label{fig:pipeline}
\end{figure*}

To address this gap, we present a proof-of-concept investigation of a persona-aware approach to LLM support (Figure~\ref{fig:pipeline}). Our contributions are threefold: \textbf{(i)} a theoretically grounded four-persona typology of self-stigma in online substance-use communities, derived via latent profile analysis (LPA) and validated against held-out behavioral features; \textbf{(ii)} a cold-start recovery evaluation demonstrating that sequential Bayesian and recurrent neural classifiers substantially outperform batch and few-shot LLM baselines under limited posting histories; and \textbf{(iii)} an expert evaluation across three contemporary LLMs revealing that persona-conditioned responses produce design-aligned content shifts but are holistically less preferred than a generic-empathy baseline.



\section{Related Works}\label{sec:background}

\noindent\textbf{Persona Archetypes From Social Media.}
Computational construction of user typologies from social media has progressed from frequency-based profiling \citep{davisArchetypeBasedModelingSearch2019, valdezComputationalAnalysesIdentify2022} to embedding-driven clustering \citep{lowNaturalLanguageProcessing2020, kimUnderstandingMentalHealth2023, sanchezrodriguezUnderstandingMentalHealth2026}. Within substance use, recent work clusters LLM-extracted stigma features \citep{wangIdentifyingStigmaPhenotypes2025} and operationalizes stigma mechanisms \citep{roeslerTheoryinformedDeepLearning2024}. Person-centered methods like LPA offer probabilistic class assignments \citep{NylundGibson2018TenFA} but are rarely applied to text-derived features. While \citet{lizzio-wilsonWhatCouldBe2025} fit an LPA on survey measures and linguistically characterized the profiles, we extend this by deriving personas from theory-grounded stigma indicators in text.

\noindent\textbf{Cold-start Classification with Mental Health Applications.}
User-level classification from limited history is central to the early-risk detection paradigm \citep{losadaERISK2017CLEF2017}. Sequential evidence accumulation \citep{burdissoTextClassificationFramework2019} and incremental embedding aggregation \citep{lianIncrementalUserEmbedding2022} effectively support early decision-making, with performance saturating rapidly as history accumulates \citep{dalalInvestigationDataRequirements2022a}. However, mental health classifiers often struggle to generalize across populations \citep{harrigianModelsMentalHealth2020}. Furthermore, multi-class classification has largely relied on full posting histories and disorder-level labels \citep{cohanSMHDLargeScaleResource2018, sanchezrodriguezUnderstandingMentalHealth2026}, leaving the cold-start recovery of latent within-condition subtypes unexplored.

\noindent\textbf{Persona-Conditioned Empathetic Response.}
Persona-conditioned empathetic dialogue has evolved from foundational datasets \citep{zhongPersonaBasedEmpatheticConversational2020} and computational frameworks \citep{sharma-etal-2020-computational}, through implicit persona inference \citep{choPersonalizedDialogueGenerator2022}, to LLM generation grounded in therapeutic strategies \citep{qianHarnessingPowerLarge2023, sharmaCognitiveReframingNegative2023, aghakhani-etal-2025-conversation}. Recent models condition responses on demographic \citep{wuPersonasTalksRevisiting2025}, personality \citep{weiMECoTMarkovEmotional2025}, or clinical profiles \citep{wangTalkDepClinicallyGrounded2025}, often utilizing chain-of-thought (CoT) prompting \citep{wangCueCoTChainofthoughtPrompting2023}. Concurrently, critical reviews urge that generative mental health chatbots must be anchored in established behavior-change theories \citep{heoUseEvaluationBehavior2026, Roshanaei2026Talk}, and practitioner-informed evaluations have documented systematic LLM failure modes (e.g., performative empathy, disregard for user context) that persist across models and prompting practices \citep{iftikharHowLLMCounselors2025, mooreLLMsTherapist2025}. To bridge these domains, our pipeline uniquely integrates text-derived persona construction, cold-start recovery, and theory-grounded empathetic generation to provide tailored support for latent behavioral subtypes.
\section{Data and Features}\label{sec:data}
Our data, annotation pipeline, and indicator taxonomy are drawn from prior work \citep{bouzoubaa2026cognitiveaffectivebehavioralexpression} that developed a theoretically grounded codebook ~\citep{corrigan2006self,corrigan2009whytry,linkModifiedLabelingTheory1989,luomaSelfStigmaSubstanceAbuse2013} for self-stigma expression. The corpus includes 72,117 Reddit posts from 1,660 users sustaining engagement with substance-use communities (e.g., \textit{r/opiates}, \textit{r/Stims}) between 2006 and 2025. Posts were annotated for self-stigma presence and ten indicators across three theoretical domains: cognitive (\emph{self-labeling}, \emph{pessimism}, \emph{deservingness/worthlessness}), affective (\emph{guilt}, \emph{shame}, \emph{despair}), and behavioral (\emph{concealment}, \emph{desire to quit}, \emph{anticipated rejection}, \emph{ambivalence}). Of the 72,117 posts, 3,838 (5.3\%) expressed self-stigma. These originated from 1,228 users (74\% of the sample), whom we term \emph{expressors}. The user served as the unit of analysis.

We computed a four-block user-level feature matrix for all expressors (full extraction details and feature list in Appendix \ref{app:features}):

\noindent\textbf{Self-Stigma Content:} Indicator rates (proportion of self-stigma posts expressing an indicator), excerpt proportions, indicator diversity, and co-occurrence density.   

\noindent\textbf{Temporal Features:} Time-dependent variables (e.g., pessimism slope) for users with $\geq$3 self-stigma posts, reserved for external validation.

\noindent\textbf{Linguistic Style: } User-level means for 96 LIWC-22 features \citep{Boyd2022} and five structural text metrics (e.g.,readability \cite{Flesch1948ANR}).

\noindent\textbf{Engagement Metadata: } Post count, posting frequency, subreddit diversity, active timeline, reserved for external validation.

\section{Phase 1: Personas of Self-Stigma}\label{sec:phase1}
We applied LPA \citep{williams2016latent} following a standard three-step framework \citep{asparouhov2014auxiliary}: fit an unconditional base model on the \textit{Self-Stigma Content} features, screen candidate covariates for profile differentiation, and fit an augmented model with empirically selected features. This prevents covariates from distorting the base profile structure.

\subsection{Model Construction}\label{sec:phase1-step1}

We first reduced the 22 \textit{Self-Stigma Content} features to 15 to remove severe collinearity ($|r|>0.70$; details in Appendix~\ref{app:lpa-config}). Initial LPA on these 15 features using \texttt{mclust} \citep{Scrucca2016} indicated a five-profile base structure. To incorporate linguistic style without circularity, we screened 101 candidate features from the \textit{Linguistic Style} category against these base profiles. We selected the top eight LIWC features based on Kruskal-Wallis effect sizes ($\eta^2 > 0.01$; \citealp{cohen2013statistical}): \emph{fatigue}, \emph{family}, \emph{conflict}, \emph{anger}, \emph{reward}, \emph{Clout}, \emph{need}, and \emph{negative tone} (full ranking in Appendix~\ref{app:screening}). Our final augmented LPA was fit on these 23 features using a VEI covariance parameterization (Appendix~\ref{app:lpa-config}).

\subsection{Results: Persona Typology}\label{sec:phase1-results}

\begin{figure*}
    \centering
    \includegraphics[width=0.99\textwidth]{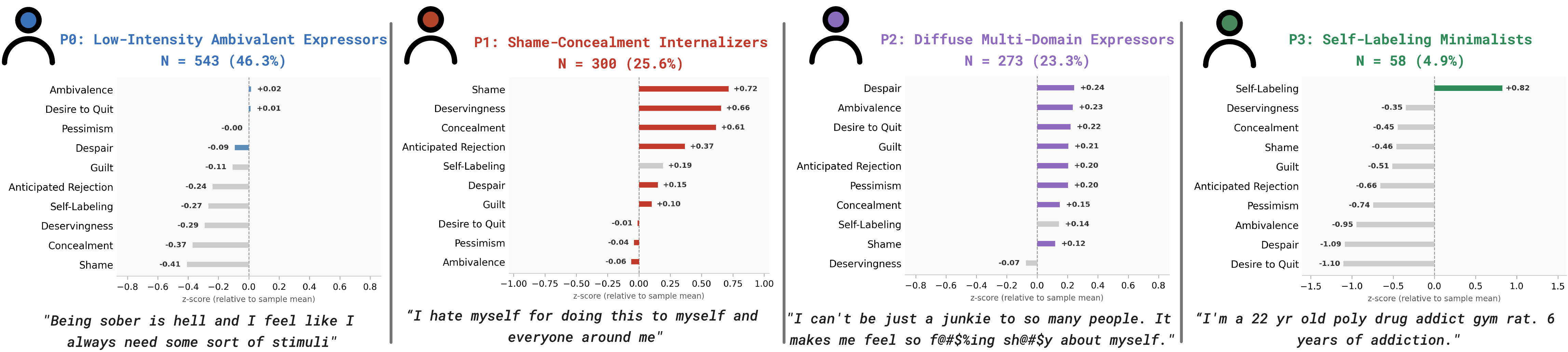}
    \caption{Four self-stigma personas identified via LPA. Each card shows the top distinguishing \textit{Self-Stigma Content} as z-scored deviations from the sample mean, with a representative paraphrased excerpt. }
    \label{fig:personas}
\end{figure*}

Model selection metrics strongly favored a four-persona solution ($k=4$). While Bayesian information criterion (BIC) showed a classical elbow at $k{=}3$, 5-fold cross-validated stability was substantially higher and more consistent for $k{=}4$ (ARI $= 0.795$, SD $= 0.104$ vs.\ $0.689$, SD $= 0.303$; Table~\ref{tab:fit-stats}). Full fit statistics across all $k$ values are reported in Appendix~\ref{app:lpa-config}. The resulting four personas (P0 to P3) are summarized in Table~\ref{tab:personas} and Figure~\ref{fig:personas}.

\noindent\textbf{Persona characterizations.}
P0 (46.3\%), the largest group, expressed self-stigma primarily through ambivalence and desire to quit with very low rates of shame and concealment. P1 (25.6\%) exhibited high shame, concealment, and deservingness alongside the highest fatigue, conflict, and anger LIWC scores. P2 (23.3\%) expressed self-stigma broadly across all ten indicator categories and featured the highest engagement volume (59\% recurrent posters). P3 (4.9\%) uniquely relied on extreme self-labeling (e.g.,``I'm a junkie'') with virtually no emotional self-stigma, distinguished linguistically by high Clout and low negative tone.Full characterizations are reported in Appendix~\ref{app:personas}.

\noindent\textbf{External Validation.}
The $k=4$ structure demonstrated strong recovery stability (five-fold CV ARI $=0.795$, SD $= 0.104$; \citep{steinley2004properties}). To confirm these text-derived personas capture distinct behavioral and engagement patterns rather than modeling artifacts, we tested them against 115 held-out features from the \textit{Temporal}, \textit{Linguistic Style}, and \textit{Engagement Metadata} categories. Following Benjamini-Hochberg FDR correction ($q < 0.05$; \citep{Benjamini1995}), 73 of these 115 features (63.5\%) significantly differentiated the four personas(full breakdown in Appendix~\ref{app:validation}). Robustness analyses are also reported in Appendix~\ref{app:sensitivity}.

\begin{table}[t]
\centering
\small
\setlength{\tabcolsep}{3pt}
\begin{tabular}{@{}p{0.05\columnwidth}p{0.28\columnwidth}p{0.14\columnwidth}p{0.48\columnwidth}@{}}
\toprule
& \textbf{Persona} & \textbf{N (\%)} & \textbf{Defining pattern} \\
\midrule
P0 & Low-Intensity Ambivalent Expressors & 543 (46.3) & Moderate despair, desire to quit, ambivalence; very low shame and concealment; narrow expression (diversity = 3.2) \\
P1 & Shame-Concealment Internalizers & 300 (25.6) & Highest shame, concealment, deservingness; elevated guilt and anticipated rejection; broad co-occurrence \\
P2 & Diffuse Multi-Domain Expressors & 273 (23.3) & Elevated across all indicators; highest diversity (7.6); highest engagement (mean 86 posts, 7.5 SS posts) \\
P3 & Self-Labeling Minimalists & 58 (4.9) & Extreme self-labeling (0.91) with near-zero on all other indicators; lowest diversity (1.2) \\
\bottomrule
\end{tabular}
\caption{Four self-stigma personas derived from LPA on 23 features (15 Block~A + 8 LIWC). P3 falls marginally below the 5\% viability threshold ($N{=}58$) but was retained due to its extreme profile distinctiveness (self-labeling rate $=0.91$, all other indicators ${<}0.02$) and the superior cross-validation stability of $k{=}4$ vs.\ $k{=}3$.}
\label{tab:personas}
\end{table}

\section{Phase 2: Early Persona Recovery}\label{sec:phase2}
 
Phase~1 established that four self-stigma personas exist; Phase~2 asks whether they can be recovered from limited data. In a deployment context, the system observes a user's posts as they arrive and must decide when it has seen enough to act. 
 
\subsection{Features and Baselines}\label{sec:phase2-features}
 
We adapted our features for the sequential setting by dropping meta-features that require complete posting history. Whether to compute statistics over all posts (\textsc{all-posts}) or restrict them strictly to self-stigma posts  (\textsc{ss-only}) is treated as an experimental condition. 
 
To evaluate sequential classification, we structured our input vectors into four distinct feature blocks extracted from a user's first $n$ posts ($n \in \{1, 3, 5, 10, 20, 30\}$): \textbf{Block~A} (20 features): self-stigma indicator rates and excerpt count means; \textbf{Block~B} (8 features): Phase~1 LIWC metrics; \textbf{Block~C} (2 features): self-stigma post proportion and log word count; and \textbf{Block~D} (10 features): mean normalized character offset of indicator's excerpts (i.e., discourse structure). This yields a 40-dimensional vector at each truncation point.

\noindent\textbf{Oracle and Encoder Baselines.} Oracle classifiers evaluated on the full posting history provide a performance ceiling. We evaluated six classifiers under stratified 5-fold cross-validation. Hyperparameters and per-fold splits are reported in Appendix~\ref{app:oracle-baselines}. The post-aggregated oracle (all posts, all 40 features) achieved macro-F1 $= 0.619$ (HGBT). When Blocks~A and~D were restricted to self-stigma posts only, the ceiling raised to $0.861$ (SVM-RBF).  In contrast, frozen user-level pre-trained encoders performed poorly: MentalRoBERTa \citep{ji2022mentalbert} achieved $0.382$, and ModernBERT \citep{modernbert} achieved $0.366$. This 23-point gap confirms that persona distinctions rely on structured, construct-specific signals that dense embeddings fail to capture. The majority baseline (predict P0) yields macro-F1 $\approx 0.158$. 
 
 
 
\subsection{Sequential Classification Experiments}\label{sec:phase2-experiments}

We evaluated four model classes at each truncation point $n$.
 
\noindent\textbf{Batch Classification.} As a non-sequential reference, batch classifiers aggregate the first $n$ posts into a single 40-dimensional vector. A $2^4$ factorial feature ablation identifies ABD+SVM as the strongest combination at $n{=}30$ ($0.476$) (see Appendix~\ref{app:ablation}). Block~D contributed the largest main effect ($+0.062$), while Block~C was unhelpful ($-0.008$).

\noindent\textbf{Bayesian accumulator.} Following the incremental confidence framework of \citet{burdissoTextClassificationFramework2019}, this model processes posts sequentially, maintaining a posterior $P(k \mid \mathbf{x}_{1:t})$ over the four personas after $t$ posts. At each new post, three steps run in order: (i) carry forward the previous posterior $P(k \mid \mathbf{x}_{1:t{-}1})$ as the new prior; (ii) compute the per-post likelihood $P(\mathbf{x}_t \mid k)$ as the product of 10 Bernoulli indicator likelihoods; (iii) apply Bayes' rule and renormalize. In log space:
 
\begin{align}\label{eq:accumulator}
\log P(k \mid \mathbf{x}_{1:t}) &= \log P(k \mid \mathbf{x}_{1:t{-}1}) \nonumber \\
&\quad + \textstyle\sum_{j=1}^{10} \log P(\text{ind}_j^{(t)} \mid k)
\end{align}
 
\noindent The initial prior reflects empirical class proportions, and likelihoods are estimated with Laplace smoothing ($\alpha{=}1$). Because self-stigma posts carry more indicators, they naturally drive larger posterior updates (justification in Appendix~\ref{app:accumulator-detail}). We tested three variants: \textbf{M1 (uniform)} estimates a single likelihood from all training posts; \textbf{M2 (SS-conditional)} utilizes separate likelihood parameters depending on a post's self-stigma flag; and \textbf{M3 (SS-only)} estimates likelihoods strictly from self-stigma posts, skipping non-self-stigma posts entirely so they produce no update.
 

\noindent\textbf{Sequential Neural Classifier (SNPC).} Because the Bayesian accumulator assumes conditional independence, it cannot detect cross-post temporal patterns (e.g., escalating shame, desisting concealment). To capture these dynamics, we trained unidirectional GRU and LSTM models that process the first $n$ posts as a sequence, passing the final hidden state through a softmax head over the four personas. We evaluated a $2 \times 2 \times 15 \times 6$ factorial grid across architecture (GRU, LSTM), input filtering (\textsc{ss-only}, \textsc{all-posts}), feature subsets (all 15 non-empty combinations of \{A, B, C, D\}), and truncation $n$. Each configuration was evaluated using stratified 5-fold cross-validation, with an 80/20 train-validation split for early stopping. We also tested hidden-size sensitivity ($h \in \{32, 64, 128\}$) on the optimal configuration, finding both architecture and hidden size to be largely inert (max $\Delta < 2.3$pp; training details in Appendix~\ref{app:snpc-config}).

\noindent\textbf{Few-shot LLM with CoT.} As a final reference, we prompted GPT5.4-Mini (\texttt{gpt-5.4-mini-2026-03-17)} and GPT-5.4-Nano (\texttt{gpt-5.4-nano-2026-03-17}) \cite{openaiGPT54_2026} with the user's first $n$ posts, two example posts per persona ($4 \times 2 = 8$ shots total), and a brief description, asking the model to assign a label via CoT reasoning. We test both \textsc{ss-only} and \textsc{all-posts} input conditions; the full prompt and per-fold results are in Appendix~\ref{app:llm}.




\subsection{Classification Results}\label{sec:phase2-results}
 
Table~\ref{tab:master-results} consolidates all Phase~2 experiments. We highlight three key findings.

\noindent\textbf{Sequential structure outperformed feature breadth.}
Aggregating posts into a single vector destroys the per-post indicator patterns that sequential methods exploit. The simple Bayesian accumulator (M3), utilizing only 10 binary indicators, reached $0.566$ at $n{=}30$. This surpassed the best 40-feature batch classifier (ABD+SVM) by $+9$ percentage points (pp). The SNPC (GRU, Blocks A+D, \textsc{ss-only}, $h{=}128$) gained an additional $+17.1$pp over M3 to reach $0.737$, coming within $0.124$ of the SS-conditional oracle. Furthermore, while the linguistic Block B helped batch models, it harmed sequential models (Appendices~\ref{app:ablation} and \ref{app:snpc-ablation}), indicating that post-level linguistic variation introduces noise that recurrent states cannot filter.
 
\noindent\textbf{Non-SS posts help early, hurt later.}
Non-self-stigma posts provide early context but eventually dilute the signal. Among Bayesian variants, M2 (SS-conditional) performed best at early stages $n{\leq}5$ when users lacked explicit self-stigma posts. However, by $n{=}10$, M3 (SS-only) surpassed, widening the gap monotonically ($0.566$ vs.\ $0.520$ at $n{=}30$). For the SNPC models, the \textsc{ss-only} condition dominated \textsc{all-posts} at every truncation point ($\Delta = +0.06$ at $n{=}1$, $+0.19$ at $n{=}30$). In stark contrast, LLM performance trajectories remained flat or declined as history length increased (e.g., GPT-5.4-Nano dropping from $0.540$ to $0.314$). 
 
\noindent\textbf{Per-persona patterns.}
Different personas favored different methods. P3 (Self-Labeling Minimalist) was suited for the M3 Bayesian accumulator, reaching F1 $= 0.770$ at $n{=}30$ and dwarfing the batch classifier ($0.250$). Conversely, P1 (Shame-Concealment Internalizer) saw the largest gains from neural modeling, jumping from $0.337$ under M3 to $0.739$ under SNPC. P2 (Diffuse Multi-Domain) remained the most difficult group to classify early, showing the largest gap ($0.196$) below its theoretical oracle ceiling. 
 

 
\begin{figure*}[t]
\centering
\includegraphics[width=.85\textwidth]{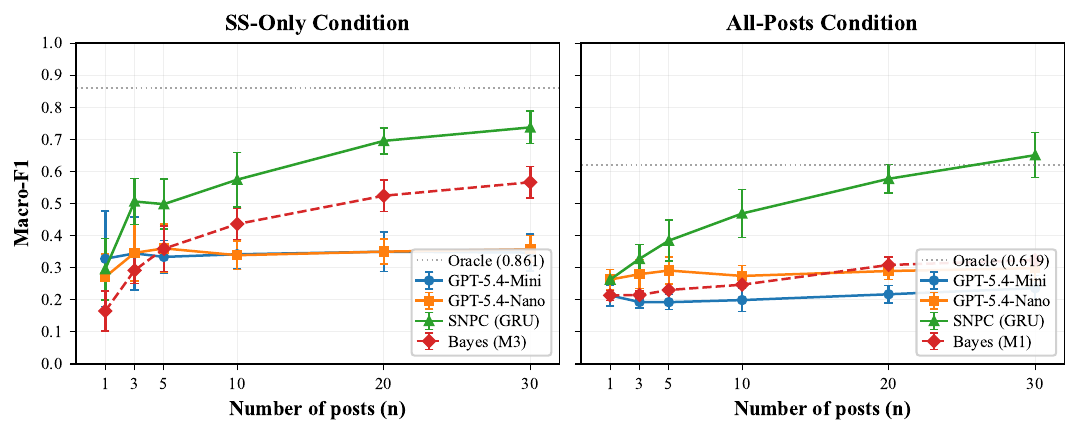}
\caption{Persona recovery (macro-F1) vs. posting history length. \textbf{SS-Only} (left): only self-stigma posts are processed; oracle ceiling is the SS-conditional oracle (0.861). \textbf{All-Posts} (right): all posts are processed; oracle ceiling is the post-aggregated oracle (0.619). SNPC = GRU with Blocks A+D under SS-Only, A+C under All-Posts (best per condition). Bayes = M3 (SS-Only) and M1 (All-Posts). Error bars show ±1 SD across 5 stratified folds.}
\label{fig:recovery_curve}
\end{figure*}
 

 \begin{table*}[t]
\centering
\small
\renewcommand{\arraystretch}{1.15}

\begin{tabular}{llccccccc}
\toprule
& & \multicolumn{6}{c}{\textbf{Truncation} $n$} & \textbf{All} \\
\cmidrule(lr){3-8}
\textbf{Method} & \textbf{Features} & $1$ & $3$ & $5$ & $10$ & $20$ & $30$ & posts \\
\midrule
Majority (predict P0) & --- & .158 & .158 & .158 & .155 & .150 & .148 & --- \\
\addlinespace
\multicolumn{9}{l}{\textit{Bayesian accumulator (sequential, 10 binary indicators)}} \\
\quad M1: uniform        & Ind & .214 & .215 & .231 & .247 & .309 & .324 & --- \\
\quad M2: SS-conditional & Ind & .234 & .314 & .369 & .418 & .487 & .520 & --- \\
\quad M3: SS-only        & Ind & .165 & .292 & .359 & .436 & .524 & .566 & --- \\
\addlinespace
\multicolumn{9}{l}{\textit{Sequential neural classifier (GRU, A+D, SS-only, $h{=}128$)}} \\
\quad SNPC               & 30 & \textbf{.296} & \textbf{.506} & \textbf{.498} & \textbf{.574} & \textbf{.695} & \textbf{.737} & --- \\
\addlinespace
\multicolumn{9}{l}{\textit{Batch classifiers (aggregated, 40 features)}} \\
\quad Best combo$^a$            & 40f          & .273 & .297 & .315 & .344 & .400 & .476 & --- \\
\quad Best combo (SS-cond.)$^b$ & 40f$_{\rm SS}$ & .175 & .304 & .344 & .395 & .469 & .495 & --- \\
\addlinespace
\multicolumn{9}{l}{\textit{Few-shot LLM (gpt-5.4-nano, CoT prompting)$^c$}} \\
\quad SS-only & raw text & .271 & .346 & .360 & .339 & .350 & .358 & --- \\
\quad All posts & raw text & .263 & .280 & .291 & .274 & .290 & .299 & --- \\
\addlinespace
\multicolumn{9}{l}{\textit{Oracle baselines (full posting history)}} \\
\quad HGBT (post-aggregated)   & 40            & --- & --- & --- & --- & --- & --- & .619 \\
\quad SVM-RBF (SS-conditional) & 40$_{\rm SS}$ & --- & --- & --- & --- & --- & --- & \textbf{.861} \\
\quad MentalRoBERTa (frozen)   & 2{,}304       & --- & --- & --- & --- & --- & --- & .382 \\
\quad ModernBERT (frozen)      & 2{,}304       & --- & --- & --- & --- & --- & --- & .366 \\
\midrule
$N$ (M1/M2/M3, batch)$^d$ & & 1{,}174 & 1{,}164 & 1{,}098 & 944 & 705 & 531 &  1{,}174 \\
$N$ (SNPC, SS-only)$^e$   & & 125  & 335  & 462  & 696 & 923 & 1{,}023 & --- \\
\bottomrule
\end{tabular}
\vspace{2pt}
 
{\footnotesize
$^a$ACD+RF ($n{\leq}10$), ABD+SVM ($n{\geq}20$). $^b$ABCD+SVM. $^c$Few-shot prompting with chain-of-thought reasoning over the user's first $n$ posts; full prompt in Appendix~\ref{app:llm}. $^d$Strict truncation: users with $\geq n$ total posts. $^e$Flexible truncation: users with $\geq n$ self-stigma posts; first $n$ SS posts used. Blocks: A = rates + counts (20); B = LIWC (8); C = SS prop.\ + word count (2); D = positions (10). Ind = 10 binary indicators per post.Per-persona, ablation, and per-fold tables in Appendices~\ref{app:per-persona}, \ref{app:ablation}, and \ref{app:snpc-ablation}.
}
\caption{Master results across all Phase~2 experiments. Mean macro-F1 across 5 stratified folds. The \textbf{All} column reports oracles using the complete posting history; $n{=}1$--$30$ columns report sequential and batch methods at successive truncation points. Best non-Oracle result per column in \textbf{bold}. $N$ rows show the evaluation cohort size at each $n$ (cohorts differ across truncation rules; see footnote).}
\label{tab:master-results}
\end{table*}

\section{Phase 3: Persona-Aware Chat}\label{sec:phase3}
Phases~1 and~2 established that four self-stigma personas exist and \textit{can} be recovered from limited posting histories. Phase~3 asks: Do persona-conditioned LLM responses produce more appropriate support than persona-neutral responses? We frame this as a deployment comparison, a persona-aware system against a naive supportive baseline, and evaluated by domain experts in clinical psychology and public health.

We composed 24 first-person, paraphrased stimuli, six per persona. Each stimulus was paired with two response conditions. The \textbf{matched} condition uses a persona-specific system using rubric-based generation \citep{hashemi-etal-2024-llm}, derived from the Phase~1 profile characterizations (\S\ref{sec:phase1}), three persona-matched few-shot examples, and a shared clinical safety floor \citep{stadeLargeLanguageModels2024}. The \textbf{neutral} condition uses a generic supportive system prompt with no demonstrations and the same safety floor. We generated responses using three contemporary LLMs spanning three providers: \texttt{gpt-5.4-mini}, \texttt{gemini-3-flash}\citep{gemini3-flash-2025}, and \texttt{llama-4-scout} \citep{llama4-scout-2025} - totaling 144 responses (24~stimuli $\times$ 3~models $\times$ 2~conditions). Generation parameters and output parsing details are in Appendix \ref{app:phase3-generation}.

\subsection{Expert Evaluation}
\label{sec:phase3-raters}
We recruited eight domain experts in clinical psychology and public health with 5--40 years of experience working with PWUD. The 144 responses were organized into 72 stimulus--model pairs, each presented on a single page, with the matched and neutral responses shown side-by-side under A/B labels. Raters were blind to the condition; each pair was rated by two experts.

For each response, raters completed 10 five-point Likert items (including two persona-specific), two multiple-choice descriptive items, and one A/B forced-choice item. The Likert set combined generic support-quality measures \citep{kumarWhenLargeLanguage2026, sharma-etal-2020-computational}, cross-cutting stigma-sensitive items (e.g., person-first language \cite{NIDA_WordsMatter}, non-performative empathy and response specificity \citep{iftikharHowLLMCounselors2025}), and persona-specific items drawn from the matched persona's rubric (Appendix \ref{app:phase3-items}).

\noindent\textbf{Analysis. } We finalized the analysis protocol before examining the evaluation results (Appendix~\ref{app:phase3-analysis}). The forced-choice item was modeled with an ordinal mixed-effects model. Per-item ratings were modeled using linear mixed-effects (rater and stimulus random effects), with multiplicity controlled across 19 tests via Benjamini--Hochberg ($q = .05$). Descriptive codes (advice category, unsignaled crisis-line referrals) were analyzed using Fisher's exact test and a binomial generalized linear mixed model (GLMM). 

\subsection{Evaluation Results}
\label{sec:phase3-results}


Across all three frontier LLMs (Gemini-3-Flash, GPT-5.4-mini, and Llama-4-Scout), matched responses behaved consistently, and the LLM $\times$ condition interaction was null (LRT $\chi^2(2) = 0.87$, $p = .65$). Item-level effects matched the persona-design predictions for three of four personas (Table~\ref{tab:phase3-per-persona}). P0 stimuli rewarded matched responses for holding tension without resolving (P0-1 $\beta = +0.56$, $q = .012$); P2 stimuli rewarded selective engagement and tolerance of unresolved distress (P2-1 $\beta = +0.79$, $q = .002$; P2-2 $\beta = +1.18$, $q < .001$). P1 stimuli rewarded suppression of elaboration probing (G1 reverse-coded; $\beta = -1.42$, $q < .001$). The universal Non-Performative Empathy item also favored matched responses ($\beta = +0.53$, $q < .001$).
 
At the content level, matched responses produced shifts uniformly across personas. Unsignaled crisis-line referrals (CRR) dropped from 48.6\% (neutral) to 1.4\% (matched), pooled across personas (binomial GLMM OR $= 0.002$, 95\% CI $[0.0003, 0.018]$, one-sided $p < 10^{-8}$; $\alpha = 0.80$). The ``No advice'' descriptive code rose from 16\% to 52\% under matched generation (per-persona Fisher's exact $p \leq .03$). Matched CRR rates were within two percentage points of each other across models (0\% Llama, 2\% Gemini, 2\% GPT).
 
Holistic expert judgment diverged from the item-level results. On the forced-choice item, raters preferred neutral responses 3.4:1 in raw votes (ordinal CLMM $\mu = -1.26$, 95\% CI $[-2.15, -0.37]$). A condition $\times$ persona interaction (LRT $\chi^2(3) = 15.17$, $p = .002$) localized this composite-level effect to P3 ($\beta_{\text{P3}} = -0.48$, 95\% CI $[-0.71, -0.24]$). P1 and P2 modestly favored matched ($\beta = +0.10$ and $+0.12$), while P0 was nearly null. Universal items measuring generic empathy (e.g., emotional validation (G2), focus on user (G3), and engagement with concerns (G4)) tracked the holistic preference, suggesting global judgment indexes generic empathy that is separable from the persona-specific qualities the rubrics were designed to enforce. In summary, while persona-matched responses successfully achieved their targeted behavioral shifts, raters holistically preferred the neutral baseline. 

\begin{table*}[t]
\centering
\small
\setlength{\tabcolsep}{5pt}
\renewcommand{\arraystretch}{1.15}
\resizebox{\textwidth}{!}{%
\begin{tabular}{@{}l c c c l r@{\hspace{2pt}}l@{}}
\toprule
\textbf{Persona} & $n$ & \textbf{CRR\% (M/N)} & \textbf{No-advice\% (M/N)} & \textbf{Largest persona-aligned item} & \multicolumn{2}{c}{\textbf{Composite $\beta$ [95\% CI]}} \\
\midrule
P0 Quiet Reckoner       & 72 & 0 / 72  & 76 / 8  & P0-1 holds tension $\mathbf{+0.56}^{*}$            & $-0.12$           & $[-0.39, +0.15]$ \\
P1 Hidden Self-Critic   & 72 & 0 / 31  & 89 / 0  & G1 fewer probes $\mathbf{-1.42}^{***}$             & $\mathbf{+0.10}$  & $[-0.11, +0.32]$ \\
P2 Diffuse Witness      & 68 & 6 / 56  & 76 / 18 & P2-2 tolerates distress $\mathbf{+1.18}^{***}$     & $\mathbf{+0.12}$  & $[-0.12, +0.35]$ \\
P3 Identity-Integrated  & 72 & 0 / 36  & 56 / 22 & P3-1 respects self-desc.\ $+0.33$ (ns)             & $-0.48^{*}$       & $[-0.71, -0.24]$ \\
\bottomrule
\end{tabular}
}
\caption{Per-persona Phase~3 results. CRR\%: unsignaled crisis-line referral rate. No-advice\%: ``No advice'' coding rate. M = matched, N = neutral. \textbf{Largest persona-aligned item}: highest-magnitude effect among items whose direction was specified by the persona design (G1 reverse-coded for P1). Composite $\beta$ bold = matched-favoring. $^{*}q<.05$, $^{**}q<.01$, $^{***}q<.001$ (BH-FDR; full per-item table in Appendix~\ref{app:phase3-peritem}).}

\label{tab:phase3-per-persona}
\end{table*}

\section{Discussion \& Conclusion}\label{sec:discussion}

Self-stigma among PWUD is not a monolith. It manifests in four empirically recoverable profiles, each tracing a different theoretical mechanism: P1 the shame, concealment, and anticipated rejection of internalized stigma \citep{livingstonCorrelatesConsequencesInternalized2010, linkModifiedLabelingTheory1989}, processed in this population through relational and conflict-laden language rather than isolated self-evaluation; P0 the ambivalence and desire to quit of Corrigan's ``why-try'' tension \citep{corrigan2009whytry} without affective collapse; P3 self-labeling without emotional self-stigma, consistent with community vernacular rather than internalization; and P2 self-stigma distributed pervasively across posting history rather than localized to any single domain. Each pattern implies a different appropriate response. Telehealth has expanded rapidly as a delivery model for OUD treatment \citep{calesCOVID19PandemicOpioid2022, georgiadisTelemedicationOpioidUse2023}. GenAI offers a scalable avenue for just-in-time persona-conditioned support, provided the typology can be recovered in real time. Phase 2 establishes that it can: a sequential neural classifier achieves usable accuracy from as few as 30 posts, making deployment operationally feasible. The risk lies on the response side. Frontier LLMs produce performative empathy and inappropriate clinical probing in mental health contexts \citep{iftikharHowLLMCounselors2025}, formulaic crisis routing on inputs that do not signal active crisis \citep{arnaiz2025between}, and measurable stigma toward substance-use populations alongside sycophantic reinforcement of distorted self-narratives, with these failures persisting across model scales and under best-practice system prompting \citep{mooreLLMsTherapist2025}. Our proof-of-concept addresses this with persona-matched responses calibrated to each self-stigma profile and constrained by a shared clinical safety floor \citep{stadeLargeLanguageModels2024}.

Across three frontier LLMs, matched responses produced their design-intended behavioral signatures uniformly and at the item level: holding tension for P0 ($\beta{=}{+}0.56$), suppressing elaboration probing for P1 ($\beta{=}{-}1.42$ on G1), and tolerating unresolved distress for P2 ($\beta{=}{+}1.18$). At the content level, matched responses dropped unsignaled crisis-line referrals from $48.6\%$ to $1.4\%$ and raised the ``no advice'' rate from $4.2\%$ to $62.7\%$, behaviors consistent with harm reduction's principle of meeting users where they are \citep{hawkHarmReductionPrinciples2017a} and motivational interviewing's caution against the ``righting reflex'' of premature advice-giving \citep{miller2012motivational}. Both findings address LLM-counselor failure modes documented in recent practitioner-informed evaluations \citep{iftikharHowLLMCounselors2025, arnaiz2025between}. Yet on the forced-choice item, raters preferred the persona-agnostic neutral baseline 3.4:1, with the gap sharpest for P3. The neutral baseline routinely dispensed the unsolicited advice and unsignaled crisis routing that recent literature flags as harms in stigmatized populations \citep{iftikharHowLLMCounselors2025, arnaiz2025between}.

Expert preference for responses that violate documented best practices reflects an anchoring effect to a conventional empathy template. Universal items measuring emotional validation, focus on the user, and engagement with concerns tracked the holistic preference, while persona-specific items tracked design intent. Holistic judgment thus indexes a generic empathy template separable from the persona-aligned qualities the rubric was built to capture. Evidence from user-level analyses of LLM-based emotional support points in the same direction: engagement and positive sentiment are predicted by narrated outcomes, trust, and response quality rather than emotional bond, while companionship-oriented use correlates with reported risks such as dependence and symptom escalation \citep{aghakhani2026like}. What matched responses deliberately withhold (performative affirmation, unsignaled crisis routing, comprehensive reflection of every distressing detail, person-first rewriting of identity language) is precisely what generic-empathy heuristics reward, and what non-directive harm-reduction practice asks providers to forgo. P3 makes the gap starkest, with no reframing of ``junkie,'' no pivot to change-talk, and no emotional escalation around content the user delivered flatly. 

Persona typology is real, recoverable from limited posting histories, and behaviorally tractable as a conditioning target across three contemporary LLMs. The work ahead must reconcile how to evaluate models that are clinically sound but stylistically unconventional: rubrics that decompose generic supportive-dialogue quality from population-aligned practice so the two can be tracked separately, multi-turn evaluation against user-experienced outcomes such as engagement, return, and self-reported felt-heardness \citep{yinAICanHelp2024} rather than third-party preference alone, and LLM-as-judge \citep{gu2024survey} scaling calibrated against the expert ratings collected here. A just-in-time support tool for self-stigma will need to know whom it is talking to and be evaluated against the practice principles its conditioning is meant to encode, not the empathy template users expect.

\section*{Limitations}\label{sec:limits}

We note several aspects of the work that bound the generalizations it supports. The persona structure is supported by sensitivity analyses and held-out feature differentiation (Appendix~\ref{app:phase1-validity}); the $k{=}4$ solution is, however, one of several defensible model-selection outcomes, chosen over the BIC elbow at $k{=}3$ on cross-validation stability evidence (\S\ref{sec:phase1-results}), with P3 retained despite a small cluster size on profile distinctiveness. The structure should be read as a typology of \textit{expression patterns} observable in this corpus rather than as stable identity categories, and a given user's assignment may shift over time in ways the present system does not model.

Personas and the recovery classifier are derived from English-language Reddit posts in substance-use communities, and mental-health classifiers built on social-media text have been shown to degrade when applied to new populations or platforms \citep{harrigianModelsMentalHealth2020}. A production system would need to revalidate persona structure on chat data, retrain the recovery classifier on the deployment register, and budget for an upstream feature-extraction pipeline (SS detection and indicator classification per utterance) whose errors would cascade into persona inference. The SS-conditional oracle bounds remaining headroom under perfect upstream extraction but does not eliminate this cost.

Phase~3 rests on 8 expert raters. Domain-expert evaluation in this population is resource-intensive, and rubric load per response is substantial, which practically constrains panel size. The human ratings collected here can serve as a calibration set for LLM-as-judge approaches \citep{gu2024survey}, enabling larger-scale evaluation across additional models, stimuli, and personas without proportionally scaling expert cost---a natural extension of the present study.

Findings are specific to English-language PWUD active on substance-use subreddits. Reddit users are not representative of the broader PWUD population, the underlying codebook draws from a primarily Western clinical literature, and the work does not generalize to other self-stigma populations, languages, or PWUD who do not engage with online peer support.

\section*{Ethical Considerations}

PWUD are a stigmatized and structurally vulnerable population. This work is grounded in harm reduction principles \citep{hawkHarmReductionPrinciples2017a}: we aim to meet users where they are rather than to direct them toward a particular outcome, and the persona-conditioned response strategies the system supports are explicitly non-directive on substance use. We frame the system as a proof of concept that could, in principle, deliver just-in-time persona-aware support, but not as a replacement for trained professional care \citep{stadeLargeLanguageModels2024}. Real deployment would require clinical oversight, integration with established crisis-response pathways, and clear disclosure that responses are AI-generated.

The Reddit corpus consists of publicly accessible posts; users did not consent to research use, which is a thinner basis for inclusion than direct opt-in. All excerpts in the paper are paraphrased and lightly modified for privacy, and usernames, subreddit-specific identifiers, and other linking information are excluded from all published materials. The original data collection and annotation protocol were deemed exempt by the authors' IRB. 

Persona inference from posting history is a profile-construction technology that could be repurposed for surveillance, targeted advertising, or discrimination, and the personas themselves should be read as expression patterns observable in this corpus rather than essentialized categories. We release prompts, evaluation code, and analysis scripts, but do not release a deployable classifier or the trained persona model, and we encourage downstream replications to consider equivalent restrictions and to re-derive personas in their own populations. Even with a shared clinical safety floor, LLM responses to self-stigma can fail in ways that harm vulnerable users---misjudging crisis severity, imposing pathologizing framings, or reinforcing devaluation through formulaic empathy \citep{iftikharHowLLMCounselors2025, arnaiz2025between}---and a production system would require ongoing red-teaming and crisis-response routing beyond the static safety floor.

\section*{Acknowledgments}

We thank the eight domain experts whose clinical and public-health expertise made our evaluation possible, with particular gratitude to Dr. Robert Sterling, Dr. David Bennett, Abia Hashmi, Mary Lucas, and Michelle Slawinski. We are also
grateful to all the members of the online communities whose shared
experiences this work learns from; we have worked to protect their
privacy at every stage.


\bibliography{references,custom, references_fixed}

\appendix

\section{Feature Definitions}\label{app:features}

Table~\ref{tab:features} defines all features in the user-level matrix, organized by semantic category. Features marked with $\dagger$ were used in the final LPA model (Phase~1); others served as held-out variables for external validation or Phase~2 classification inputs. Features marked with * were computed but excluded before modeling due to collinearity (see \S\ref{sec:phase1-step1}).

\begin{table*}[t]
\centering
\scriptsize
\renewcommand{\arraystretch}{1.15}
\begin{tabular}{p{0.12\textwidth}p{0.24\textwidth}p{0.50\textwidth}p{0.06\textwidth}}
\toprule
\textbf{Category} & \textbf{Feature} & \textbf{Definition} & \textbf{In model} \\
\midrule
\multicolumn{4}{l}{\textit{Self-Stigma Content (N = 1,174 expressors)}} \\
\midrule
Rates & \texttt{self\_labeling\_rate} & Proportion of SS posts expressing self-labeling & $\dagger$ \\
 & \texttt{pessimism\_rate} & Proportion of SS posts expressing pessimism/self-defeatism & $\dagger$ \\
 & \texttt{deservingness\_rate} & Proportion of SS posts expressing deservingness/worthlessness & $\dagger$ \\
 & \texttt{guilt\_rate} & Proportion of SS posts expressing guilt/self-blame & $\dagger$ \\
 & \texttt{shame\_rate} & Proportion of SS posts expressing shame & $\dagger$ \\
 & \texttt{despair\_rate} & Proportion of SS posts expressing despair/hopelessness & $\dagger$ \\
 & \texttt{concealment\_rate} & Proportion of SS posts expressing concealment & $\dagger$ \\
 & \texttt{desire\_to\_quit\_rate} & Proportion of SS posts expressing desire to quit & $\dagger$ \\
 & \texttt{anticipated\_rejection\_rate} & Proportion of SS posts expressing anticipated rejection & $\dagger$ \\
 & \texttt{ambivalence\_rate} & Proportion of SS posts expressing ambivalence & $\dagger$ \\
\addlinespace
Excerpt & \texttt{self\_labeling\_excerpt\_prop} & Fraction of user's total SS excerpts that are self-labeling & $\dagger$ \\
Proportions & \texttt{shame\_excerpt\_prop} & Fraction of user's total SS excerpts that are shame & $\dagger$ \\
 & \texttt{deservingness\_excerpt\_prop} & Fraction of user's total SS excerpts that are deservingness & $\dagger$ \\
 & \texttt{pessimism\_excerpt\_prop} & Fraction of user's total SS excerpts that are pessimism & * \\
 & \texttt{guilt\_excerpt\_prop} & Fraction of user's total SS excerpts that are guilt & * \\
 & \texttt{despair\_excerpt\_prop} & Fraction of user's total SS excerpts that are despair & * \\
 & \texttt{concealment\_excerpt\_prop} & Fraction of user's total SS excerpts that are concealment & * \\
 & \texttt{desire\_to\_quit\_excerpt\_prop} & Fraction of user's total SS excerpts that are desire to quit & * \\
 & \texttt{anticipated\_rej\_excerpt\_prop} & Fraction of user's total SS excerpts that are anticipated rejection & * \\
 & \texttt{ambivalence\_excerpt\_prop} & Fraction of user's total SS excerpts that are ambivalence & * \\
\addlinespace
Meta-Features & \texttt{indicator\_diversity} & Count of distinct indicator types ever expressed (0--10) & $\dagger$ \\
 & \texttt{co\_occurrence\_density} & Proportion of SS posts with indicators from $\geq$2 domains & $\dagger$ \\
\addlinespace
Domain Rates & \texttt{cognitive\_rate} & Mean of self-labeling, pessimism, deservingness rates & * \\
 & \texttt{affective\_rate} & Mean of guilt, shame, despair rates & * \\
 & \texttt{behavioral\_rate} & Mean of concealment, desire to quit, anticipated rejection, ambivalence rates & * \\
\midrule
\multicolumn{4}{l}{\textit{Temporal Features (N = 384 users with $\geq$3 SS posts; external validation only)}} \\
\midrule
Trajectories & \texttt{emergence\_ordering} & Whether behavioral indicators emerged before, after, or simultaneously & \\
 & \texttt{state\_entropy} & Shannon entropy of the user's compositional state sequence & \\
 & \texttt{affective\_persistence\_rate} & Proportion of state transitions remaining in affective-present state & \\
 & \texttt{modal\_state} & Most frequent compositional state across the user's SS posts & \\
 & \texttt{pessimism\_trend\_slope} & GEE slope of pessimism presence across the user's SS timeline & \\
 & \texttt{trajectory\_length} & Number of SS posts (temporal sequence length) & \\
 & \texttt{ss\_timeline\_span\_days} & Days between first and last SS post & \\
 & \texttt{[indicator]\_emergence\_pos} & Normalized position (0--1) at which each indicator first appeared & \\
\midrule
\multicolumn{4}{l}{\textit{Linguistic Style (N = 1,174 expressors)}} \\
\midrule
LIWC & \texttt{liwc\_fatigue} & LIWC-22 fatigue/exhaustion category mean & $\dagger$ \\
 & \texttt{liwc\_family} & LIWC-22 family references mean & $\dagger$ \\
 & \texttt{liwc\_conflict} & LIWC-22 conflict/disagreement language mean & $\dagger$ \\
 & \texttt{liwc\_emo\_anger} & LIWC-22 anger emotion mean & $\dagger$ \\
 & \texttt{liwc\_reward} & LIWC-22 reward/motivation language mean & $\dagger$ \\
 & \texttt{liwc\_Clout} & LIWC-22 social confidence/authority composite & $\dagger$ \\
 & \texttt{liwc\_need} & LIWC-22 need/necessity language mean & $\dagger$ \\
 & \texttt{liwc\_tone\_neg} & LIWC-22 negative emotional tone composite & $\dagger$ \\
 & \texttt{liwc\_[other]} & Remaining 88 LIWC-22 categories & \\
\addlinespace
Structural & \texttt{mean\_post\_length\_words} & Mean word count across all user posts & \\
 & \texttt{post\_length\_variance} & Variance of word count across all user posts & \\
 & \texttt{question\_rate} & Proportion of user posts containing a question mark & \\
 & \texttt{mean\_flesch\_kincaid} & Mean Flesch-Kincaid grade level across all user posts & \\
 & \texttt{mean\_ttr} & Mean type-token ratio across all user posts & \\
\midrule
\multicolumn{4}{l}{\textit{Engagement Metadata (N = 1,174 expressors; external validation only)}} \\
\midrule
Metadata & \texttt{total\_post\_count} & Total number of posts by the user & \\
 & \texttt{active\_timeline\_days} & Days between first and last post (any subreddit) & \\
 & \texttt{posting\_frequency} & Posts per active day & \\
 & \texttt{subreddit\_diversity} & Number of unique subreddits posted in & \\
 & \texttt{ss\_post\_count} & Number of self-stigma posts & \\
 & \texttt{ss\_rate} & Proportion of total posts that are self-stigma & \\
 & \texttt{primary\_substance\_class} & Most frequently referenced substance (categorical) & \\
 & \texttt{substance\_diversity} & Number of distinct substance classes referenced & \\
 & \texttt{drug\_sub\_proportion} & Proportion of posts in substance-related subreddits & \\
\bottomrule
\end{tabular}
\caption{User-level feature definitions across four semantic categories. SS = self-stigma. $\dagger$ = included in the final 23-feature LPA model. * = computed but excluded during collinearity reduction ($|r| > 0.70$) or by design decision. Unlabeled features served as external validation variables. Temporal features are available only for the 384-user subset.}
\label{tab:features}
\end{table*}
\section{Phase 1 Supplementary Analyses}\label{app:phase1}

\subsection{LPA Configuration}\label{app:lpa-config}
 
\paragraph{Sample exclusion.}\label{app:exclusion}
We excluded 54 (4.4\%) of the 1,228 expressors identified upstream because their flagged posts contained zero annotated indicators, consistent with classifier false positives. This yielded a final sample of 1,174 users.
 
\paragraph{Covariance parameterization.}
We initially targeted the VEI covariance parameterization (variable volume, equal shape, diagonal orientation) to accommodate clusters of varying sizes. However, VEI failed to converge on the initial 15 \textit{Self-Stigma Content} features due to highly disparate variance scales (e.g., desire-to-quit rate variance $= 0.18$ vs.\ deservingness excerpt proportion variance $= 0.003$). We therefore fit the base model using EEI (equal volume, equal shape), which successfully converged. Once augmented with the 8 selected LIWC features, which supply additional, stabilizing covariance structure decoupled from the core indicator rates, the final 23-feature model successfully converged under VEI.
 
\paragraph{Feature screening.}\label{app:screening}
Of the 93 candidates surviving variance screening, 27 exceeded the small-effect threshold ($\eta^2 > 0.01$) across the base profiles. The top 8 by $\eta^2$, all LIWC, were selected for the augmented LPA: \emph{fatigue} ($\eta^2{=}0.026$), \emph{family} (0.020), \emph{conflict} (0.019), \emph{anger} (0.019), \emph{reward} (0.018), \emph{Clout} (0.018), \emph{need} (0.018), and \emph{negative tone} (0.016). The highest-ranked structural text feature was \texttt{subreddit\_diversity} ($\eta^2{=}0.016$, rank 11).

\begin{table*}[t]
\centering
\small
\caption{Profile enumeration fit statistics ($N = 1{,}174$; 
23 features, VEI parameterization).}
\label{tab:fit-stats}
\begin{tabular}{crrcrrrrrr}
\toprule
& & & & & \multicolumn{4}{c}{\textbf{Class distribution} $n$ (\%)} \\
\cmidrule(lr){6-9}
$k$ & \textbf{BIC} & \textbf{$\Delta$BIC} & \textbf{Entropy} & \textbf{CV ARI (SD)} & \textbf{1} & \textbf{2} &
\textbf{3} & \textbf{4} \\
\midrule
2 & $-$70,732 & --- & 0.967 & --- &
926 (78.9) & 248 (21.1) & --- & --- \\
3 & $-$68,841 & +1,891 & 0.979 & 0.689 (0.303) &
874 (74.4) & 241 (20.5) & 59 (5.0) & --- \\
\textbf{4} & \textbf{$-$67,969} & \textbf{+872} & \textbf{0.936} & \textbf{0.795 (0.104)} &
\textbf{543 (46.3)} & \textbf{300 (25.6)} & \textbf{273 (23.3)} & \textbf{58 (4.9)} \\
5 & $-$65,809 & +2,160 & 0.944 & 0.642 (0.322) &
502 (42.8) & 335 (28.5) & 240 (20.4) & 53 (4.5)$^{\dagger}$ \\
6 & $-$64,728 & +1,080 & 0.941 & 0.816 (0.152) &
374 (31.9) & 324 (27.6) & 229 (19.5) & 155 (13.2)$^{\dagger}$ \\
7 & $-$64,104 & +624  & 0.952 & 0.688 (0.212) &
327 (27.9) & 323 (27.5) & 231 (19.7) & 142 (12.1)$^{\dagger}$ \\
8 & $-$63,110 & +994  & 0.956 & 0.656 (0.069) &
332 (28.3) & 241 (20.5) & 209 (17.8) & 141 (12.0)$^{\dagger}$ \\
9 & $-$62,060 & +1,051 & 0.959 & 0.655 (0.137) &
353 (30.1) & 183 (15.6) & 175 (14.9) & 138 (11.8)$^{\dagger}$ \\
10 & $-$60,783 & +1,277 & 0.962 & 0.563 (0.094) &
348 (29.6) & 202 (17.2) & 130 (11.1) & 105 (8.9)$^{\dagger}$ \\
\bottomrule
\end{tabular}
\vspace{4pt}
\parbox{\textwidth}{\footnotesize
BIC = Bayesian information criterion (higher = better);
Entropy = normalized classification entropy;
CV ARI = 5-fold cross-validated Adjusted Rand Index.
\textbf{Bold} = selected solution. \\ $^{\dagger}$Additional profiles below $N{=}62$ (5\%) not shown.}
\end{table*}
\subsection{Robustness and Validity}\label{app:phase1-validity}

\paragraph{Sensitivity analyses. }\label{app:sensitivity}

Table~\ref{tab:sensitivity} summarizes robustness checks conducted during persona construction, confirming the stability of the final $k=4$ structure.

\begin{table*}[t]
\centering
\small
\renewcommand{\arraystretch}{1.2}
\begin{tabular}{p{0.22\textwidth}p{0.38\textwidth}p{0.30\textwidth}}
\toprule
\textbf{Analysis} & \textbf{Procedure} & \textbf{Result} \\
\midrule
\textit{Log-transform sensitivity} & Applied log1p to 14 highly skewed features; refit VEI $k{=}4$. & ARI $= 0.756$ vs.\ untransformed. Profile structure preserved. \\
\textit{Engagement proxy check} & Pearson $r$ between \texttt{indicator\_diversity} and \texttt{total\_post\_count}. & $r = 0.216$. Diversity is not merely a proxy for volume. \\
\textit{Cross-specification} & Compared custom, all-post, and SS-only LIWC extractions. & Core structural distinctions remained stable across all setups. \\
\textit{BIC elbow override} & Compared $k{=}3$ (BIC elbow) vs.\ $k{=}4$ on stability. & Selected $k{=}4$ (ARI $= 0.795$) over $k{=}3$ (ARI $= 0.689$) for stability. \\
\bottomrule
\end{tabular}
\caption{Sensitivity analyses confirming the robustness of the final $k=4$ LPA solution.}
\label{tab:sensitivity}
\end{table*}

\paragraph{Held-out feature differentiation \& Pairwise difficulty.}\label{app:validation}
Validation against held-out features showed that engagement variables (e.g., self-stigma post count, $\eta^2 = 0.38$) most strongly differentiated the profiles. Regarding classification difficulty, pairwise Cohen's $d$ revealed that P0--P3 and P1--P3 exhibit high separation on \textit{Self-Stigma Content} ($d > 0.60$) but are near-identical linguistically ($d < 0.15$), identifying them as the primary challenge for Phase 2 cold-start models.

\subsection{Full Persona Characterizations}\label{app:personas}

Table~\ref{tab:persona-full} presents detailed characterizations for each persona across self-stigma content, linguistic style, and engagement dimensions. Representative excerpts are paraphrased for privacy.

\begin{table*}[t]
\centering
\small
\renewcommand{\arraystretch}{1.25}
\begin{tabular}{p{0.13\textwidth}p{0.19\textwidth}p{0.19\textwidth}p{0.19\textwidth}p{0.19\textwidth}}
\toprule
& \textbf{P0: Low-Intensity Ambivalent} \newline N=543 (46.3\%) & \textbf{P1: Shame-Concealment Internalizers} \newline N=300 (25.6\%) & \textbf{P2: Diffuse Multi-Domain} \newline N=273 (23.3\%) & \textbf{P3: Self-Labeling Minimalists} \newline N=58 (4.9\%) \\
\midrule

\textbf{Self-Stigma Content} \newline (z-scores) &
Ambivalence (+0.02), desire to quit (+0.01); shame ($-$0.41), concealment ($-$0.37). Diversity: 3.2. &
Shame (+0.72), deservingness (+0.66), concealment (+0.61). Co-occurrence density: +0.27. &
All indicators elevated (+0.12 to +0.24); highest diversity (+1.26) and co-occurrence (+0.25). &
Self-labeling (+0.82); all others strongly negative ($-$0.35 to $-$1.10). Diversity: $-$1.35. \\
\addlinespace

\textbf{Linguistic Style} \newline (LIWC z-scores) &
Below average on all 8 LIWC features. Lowest negative tone ($-$0.1). &
Highest on all 8 LIWC features: conflict (+0.3), family (+0.3), anger (+0.2). &
Near average on most features. Lowest Clout ($-$0.2). &
Highest Clout (+0.2). Lowest negative tone ($-$0.4), conflict ($-$0.2). \\
\addlinespace

\textbf{Engagement Metadata} &
Episodic (82\%). Mean 43 total posts, 1.9 SS posts. &
Episodic (84\%). Mean 35 total posts, 2.0 SS posts. &
59\% recurrent, 25\% pervasive. Mean 86 total posts, 7.5 SS posts. &
Episodic (84.5\%). Mean 39 total posts, 1.6 SS posts. Lowest SS rate (0.075). \\
\addlinespace

\textbf{Theoretical Mapping} &
Reflects the ``why try'' dynamic \citep{corrigan2009whytry}: ambivalence and desire to quit without deep internalization. Self-stigma is acknowledged but not settled as self-knowledge. &
Maps onto internalized stigma \citep{livingstonCorrelatesConsequencesInternalized2010}: stigmatizing beliefs accepted as valid. Co-elevation of shame, concealment, and anticipated rejection aligns with modified labeling theory \citep{linkModifiedLabelingTheory1989}. &
Chronic, multi-domain expression. Self-stigma permeates engagement rather than concentrating in one domain. Low Clout indicates linguistic diffidence. &
Self-labeling without emotional internalization. May represent identity integration, community convention (``addict'' as argot), or nascent self-stigma before affective processing develops. \\
\addlinespace

\textbf{Representative Excerpts} \newline (paraphrased) &
``Living sober is pure hell, my brain is just constantly screaming for some kind of stimulus.'' \newline
``I'm dying to get high'' \newline
``Things are finally looking up for me, my whole life is actually turning around.'' &
``I absolutely f***ing hate myself for putting my family and myself through this garbage.'' \newline
``Right before I dose, the amount of shame that hits me is just crushing.'' \newline
``I've somehow tricked her into thinking I got completely clean ages ago.'' &
``I can't be just a junkie to so many people. It makes me feel so f***ing s***ty about myself.'' \newline
``I think of myself as beneath everyone else'' \newline
``Just letting myself become an addict makes me feel like I'm permanently tainted.'' &
``Young gym bro here, also a poly-addict for the last half-decade.'' \newline
``I already know I'm f***ed up'' \newline
`There's got to be other functional, low-key addicts lurking on here.'' \\

\bottomrule
\end{tabular}
\caption{Full persona characterizations across self-stigma content, linguistic style, engagement metadata, and theoretical grounding. SS = self-stigma. Excerpts are drawn from high-confidence persona members (posterior probability $\geq 0.90$) and lightly modified for anonymity.}
\label{tab:persona-full}
\end{table*}

\section{Phase 2 Supplementary Analyses}\label{app:phase2}

\subsection{Oracle Baselines}\label{app:oracle-baselines}

Table~\ref{tab:oracle-baselines} details the six oracle configurations. The post-aggregated oracle (0.619, HGBT) and SS-conditional oracle (0.861, SVM) are referenced in the main text; remaining configurations provide context.

\begin{table*}[t]
\centering
\small
\renewcommand{\arraystretch}{1.15}
\caption{Oracle baselines using complete posting histories, mean macro-F1 across 5 folds. \textit{4-class}: persona recovery among $N{=}1{,}174$ expressors. \textit{5-class}: adds 432 non-expressors as a fifth class ($N{=}1{,}606$). \textit{SS-cond.}: Blocks A and D computed with self-stigma posts as denominator. \textit{LPA constr.}: original 23 Phase~1 construction features (z-scored). \textit{MentalRoBERTa}: frozen domain-adapted encoder embeddings (2{,}304-dim mean+max+std pooling). \textit{ModernBERT}: frozen general-purpose encoder embeddings (2{,}304-dim, same pooling). Best result per column in \textbf{bold}.}
\label{tab:oracle-baselines}
\begin{tabular}{lcccccc}
\toprule
& \multicolumn{6}{c}{\textbf{Oracle variant}} \\
\cmidrule(lr){2-7}
\textbf{Classifier} & \textbf{4-class} & \textbf{5-class} & \textbf{SS-cond.} & \textbf{LPA constr.} & \textbf{MentalRoBERTa} & \textbf{ModernBERT} \\
& (40f) & (40f) & (40f$_\text{SS}$) & (23f) & (2{,}304-dim) & (2{,}304-dim) \\
\midrule
HGBT               & \textbf{.619} & \textbf{.692} & .842          & .903          & .346          & .326          \\
SVM (RBF)          & .602          & .600          & \textbf{.861} & \textbf{.951} & .381          & \textbf{.366} \\
LogisticRegression & .576          & .639          & .774          & .924          & \textbf{.382} & .338          \\
RandomForest       & .556          & .640          & .842          & .892          & .285          & .286          \\
GaussianNB         & .463          & .401          & .847          & .861          & .286          & .288          \\
MLP                & .558          & .583          & .751          & .889          & .340          & .352          \\
\bottomrule
\end{tabular}
\vspace{2pt}

{\footnotesize
5-class oracle: non-expressor recall = 1.0 (perfect separation); P3 recall = 0.243 (69\% absorbed into P0). \\
ModernBERT: \texttt{answerdotai/ModernBERT-base}; embeddings extracted under the same protocol as MentalRoBERTa.
}
\end{table*}

\subsubsection{Non-Expressor Identifiability}\label{app:5class-confusion}

Table~\ref{tab:5class-confusion} reports the 5-class oracle confusion matrix. Non-expressors are perfectly separated from all expressor personas (recall $= 1.0$, precision $= 1.0$), driven by metadata features (zero SS proportion and indicator rates). Among expressors, introducing the fifth class degrades P3 recovery: 40 of 58 P3 users (69\%) are absorbed by P0, dropping P3 recall to 0.243. The 5-class macro-F1 (0.692) exceeds the 4-class (0.619) only because the easily separated non-expressor class artificially inflates the average; per-persona expressor performance is uniformly worse.

\begin{table}[h]
\centering
\small
\caption{5-class oracle confusion matrix (HGBT, 5-fold sum). Non-expressors are perfectly separated; P3 is predominantly absorbed into P0.}
\label{tab:5class-confusion}
\begin{tabular}{lccccc}
\toprule
& \multicolumn{5}{c}{\textbf{Predicted}} \\
\cmidrule(lr){2-6}
\textbf{True} & P0 & P1 & P2 & P3 & Non-exp \\
\midrule
P0 (543)      & \textbf{442} & 49  & 40  & 12 & 0 \\
P1 (300)      & 70  & \textbf{191} & 35  & 4  & 0 \\
P2 (273)      & 50  & 27  & \textbf{196} & 0  & 0 \\
P3 (58)       & 40  & 3   & 1   & \textbf{14} & 0 \\
Non-exp (432) & 0  & 0   & 0   & 0  & \textbf{432} \\
\bottomrule
\end{tabular}
\end{table}

\subsubsection{Encoder Embedding Oracles}\label{app:mentalroberta}

To test whether dense text representations capture persona-discriminative signals, we encoded all posts with two frozen transformers: MentalRoBERTa \citep{ji2022mentalbert} (domain-adapted baseline; 125M parameters) and ModernBERT \citep{modernbert} (general-purpose baseline; 149M parameters).

For each post, we extracted the [CLS] token embedding (\texttt{max\_length}${}=512$). User representations were generated by concatenating mean, max, and standard-deviation pooling across per-post embeddings (2{,}304 dimensions). Evaluated under the same oracle protocol, MentalRoBERTa (LogisticRegression) reached macro-F1 $= 0.382$, and ModernBERT (SVM-RBF) reached $0.366$. Both fell over $23$pp below the structured HGBT oracle ($0.619$), confirming that dense embeddings fail to isolate this construct-specific signal.


\subsection{Feature Block Ablation (Batch Classifier)}\label{app:ablation}

Table~\ref{tab:ablation} reports macro-F1 for all $2^4$ block combinations at $n{=}10$ and $n{=}30$. Block~D (excerpt positions) provides the largest main effect at $n{=}30$ ($+0.062$), followed by Block~B (LIWC; $+0.044$). Block~C (SS proportion and word count) contributes nothing ($-0.008$), indicating that raw expression volume adds no discriminative value once content and linguistic style are modeled.

The best combination is ABD+SVM ($0.476$ at $n{=}30$). If Blocks~A and D are computed using only SS posts, the optimal combination shifts to ABCD+SVM ($0.495$); Block~C becomes informative because SS proportion now indexes the reliability of the indicator rates. We also include a baseline using only the 10 binary indicator rates (A$_\text{binary}$; $0.381$ at $n{=}30$) to match the Bayesian accumulator's feature space. The accumulator's $0.566$ on this identical feature space confirms that sequential processing adds $+0.185$ over batch aggregation.

\begin{table*}[t]
\centering
\small
\renewcommand{\arraystretch}{1.10}
\caption{Feature block ablation: mean macro-F1 (5-fold CV) for all $2^4$ combinations of feature blocks at $n{=}10$ and $n{=}30$. Each cell shows the best classifier; classifier identity in parentheses. Feature blocks: A = indicator rates + counts (20 features); B = LIWC (8); C = SS proportion + word count (2); D = excerpt positions (10). Best result per column in \textbf{bold}.}
\label{tab:ablation}
\begin{tabular}{ccccccc}
\toprule
\textbf{A} & \textbf{B} & \textbf{C} & \textbf{D} & \textbf{Features} & \textbf{$n{=}10$} & \textbf{$n{=}30$} \\
\midrule
\cmark & \xmark & \xmark & \xmark & 20 & .301 (RF) & .378 (LR) \\
\xmark & \cmark & \xmark & \xmark & 8 & .273 (NB) & .319 (SVM) \\
\xmark & \xmark & \cmark & \xmark & 2 & .268 (RF) & .324 (SVM) \\
\xmark & \xmark & \xmark & \cmark & 10 & .291 (SVM) & .383 (SVM) \\
\addlinespace
\cmark & \cmark & \xmark & \xmark & 28 & .317 (LR) & .425 (SVM) \\
\cmark & \xmark & \cmark & \xmark & 22 & .337 (RF) & .363 (LR) \\
\cmark & \xmark & \xmark & \cmark & 30 & .333 (HGBT) & .403 (SVM) \\
\xmark & \cmark & \cmark & \xmark & 10 & .302 (HGBT) & .361 (SVM) \\
\xmark & \cmark & \xmark & \cmark & 18 & .331 (SVM) & .462 (SVM) \\
\xmark & \xmark & \cmark & \cmark & 12 & .310 (SVM) & .402 (SVM) \\
\addlinespace
\cmark & \cmark & \cmark & \xmark & 30 & .329 (HGBT) & .417 (SVM) \\
\cmark & \cmark & \xmark & \cmark & 38 & .325 (LR) & \textbf{.476} (SVM) \\
\cmark & \xmark & \cmark & \cmark & 32 & \textbf{.344} (RF) & .400 (SVM) \\
\xmark & \cmark & \cmark & \cmark & 20 & .337 (SVM) & .459 (SVM) \\
\cmark & \cmark & \cmark & \cmark & 40 & .339 (SVM) & .466 (SVM) \\
\addlinespace
\midrule
\multicolumn{5}{l}{\textit{Additional comparison rows ($n{=}30$ only):}} & & \\
\multicolumn{5}{l}{\quad A$_\text{binary}$ (10 indicator rates only)} & --- & .381 (LR) \\
\multicolumn{5}{l}{\quad AB (28 features)} & --- & .425 (SVM) \\
\midrule
\multicolumn{7}{l}{\textbf{Block main effects} ($n{=}30$, marginalizing over all other blocks):} \\
\multicolumn{5}{l}{\quad D (excerpt positions)} & & $+.062$ \\
\multicolumn{5}{l}{\quad B (LIWC)} & & $+.044$ \\
\multicolumn{5}{l}{\quad A (indicator rates + counts)} & & $+.029$ \\
\multicolumn{5}{l}{\quad C (SS proportion + word count)} & & $-.008$ \\
\bottomrule
\end{tabular}
\vspace{2pt}

{\footnotesize
Classifier abbreviations: SVM = SVM~RBF, LR = LogisticRegression, RF = RandomForest, HGBT = HistGradientBoosting, NB = GaussianNB. Main effects computed as mean macro-F1 with block present minus absent, using best classifier per combination.
}
\end{table*}

\subsection{Bayesian Accumulator Analyses}\label{app:accumulator}

This subsection groups four analyses specific to the Bayesian accumulator: the design-choice justifications referenced from the body (\S\ref{app:accumulator-detail}), the threshold-tuning protocol (\S\ref{app:threshold-sensitivity}), posterior calibration across variants (\S\ref{app:calibration}), and stratified performance by SS-density (\S\ref{app:ss-density}).

\subsubsection{Estimation, Smoothing, and Justification}\label{app:accumulator-detail}

\paragraph{No explicit SS-weighting parameter.} An explicit upweighting term for SS posts is unnecessary. SS posts carry an average of $3.13$ indicators (median: $3$); non-SS posts contain only $1.55$ ($1$). Because each indicator contributes an additive log-likelihood term to the posterior update, SS posts organically produce roughly twice the evidence magnitude. This embedded asymmetry avoids an extra tuning parameter.

\paragraph{Restricting to binary indicators.} First, Bernoulli likelihoods are exact for binary variables but require arbitrary binning/discretization for continuous features like Block~B (LIWC). Second, including binned LIWC features would expand the parameter space from 40 to 360 parameters, increasing variance without clear benefit. Third, the batch classifier (ABD+SVM) uses the full 40-dimensional continuous vector but still underperforms the 10-feature accumulator at every truncation point, suggesting the gap lies in sequential structure, not feature breadth.

\paragraph{Smoothing and prior.} The likelihood $P(\text{ind}_j^{(t)} \mid k)$ uses Laplace smoothing ($\alpha = 1$) to prevent zero probabilities on unseen combinations. The prior at $t = 0$ is set to empirical training-fold class proportions (which match global proportions within $\pm 1.5\%$) rather than uniform, reflecting true population imbalance (Table~\ref{tab:persona-full}).

\subsubsection{Threshold Sensitivity}\label{app:threshold-sensitivity}

The accumulator's abstention mechanism requires two tuned thresholds: $\theta$ (minimum posterior probability) and $\rho$ (minimum margin between top-two posteriors). Grid search across $\theta \in [0.30, 0.90]$ and $\rho \in [0.00, 0.30]$ optimized for macro-F1 $\times$ coverage.

At $n \geq 10$, the optimal thresholds were $\theta^* = 0.30$ and $\rho^* = 0.00$ (coverage $= 1.0$). Posteriors at these truncation points are sufficiently confident that abstention provides no benefit; forcing classification yields identical results. Abstention only becomes relevant at very low $n$ or in extremely sparse populations.

\subsubsection{Calibration}\label{app:calibration}

\begin{figure}[t]
\centering
\includegraphics[width=\columnwidth]{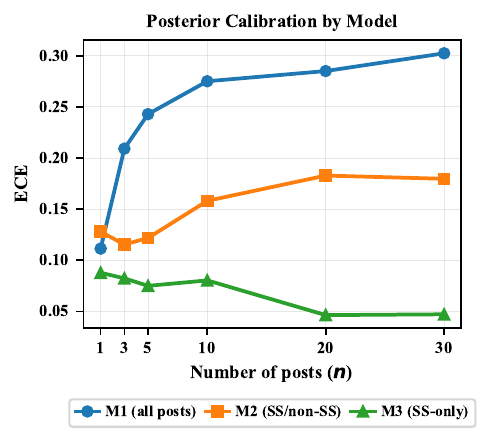}
\caption{Expected Calibration Error by model across truncation points. M3 (SS-only) improves monotonically; M1 (uniform) degrades as non-SS posts inflate posterior confidence.}
\label{fig:calibration}
\end{figure}

Expected Calibration Error (ECE) was computed using 10 equal-width bins over the maximum posterior probability. M3's ECE improves monotonically with $n$ (0.088 at $n{=}1$ $\rightarrow$ 0.047 at $n{=}30$), indicating well-calibrated posteriors. M2's ECE is moderate and stable (0.128 $\rightarrow$ 0.180). M1's ECE worsens sharply (0.111 $\rightarrow$ 0.302), consistent with non-SS posts accumulating confident but incorrect likelihood contributions (Figure~\ref{fig:calibration}). These patterns confirm that M3's posterior probabilities are trustworthy for downstream use in Phase~3 response conditioning.

\subsubsection{SS-Density Stratification}\label{app:ss-density}
At $n{=}10$, users in the highest SS-density tertile achieved macro-F1 $= 0.563$, compared to $0.470$ (medium) and $0.219$ (low). Low-density users hover near the majority baseline because the accumulator lacks indicator signals. This validates M3's design: withholding classification when evidence is sparse is strictly preferable to forcing a low-confidence prediction.

\subsection{Sequential Neural Classifier Configuration}\label{app:snpc-config}

\subsubsection{Training Configuration}



All SNPC variants employ a single-layer, unidirectional encoder (GRU or LSTM) with a final-state softmax classification head. Table~\ref{tab:snpc-hparams} summarizes the full configuration. Models were fit using stratified 5-fold CV with an internal 85/15 train/validation split for early stopping.

\begin{table}[t]
\centering
\small
\caption{SNPC training configuration. Bracketed values are placeholders to be confirmed against the released training script.}
\label{tab:snpc-hparams}
\begin{tabular}{@{}l p{0.65\columnwidth}@{}}
\toprule
\textbf{Component} & \textbf{Value} \\
\midrule
Architecture & GRU or LSTM, 1 layer, unidirectional \\
Hidden size  & 64 (primary ablation), {32, 64, 128} (sensitivity) \\
Classifier head & Linear $\rightarrow$ softmax over 4 classes \\
Optimizer    & Adam \\
Learning rate & 1e-3 \\
Weight decay & 1e-4  \\
Dropout      & 0.3 applied to recurrent output \\
Batch size   & 32 \\
Max epochs   & 100 \\
Early-stop patience & 10 epochs on val macro-F1 \\
Loss         & Cross-entropy \\
Random seed  & 42 \\
CV scheme    & 5-fold stratified; 85/15 train/val within each training fold \\
\bottomrule
\end{tabular}
\end{table}

\subsubsection{Block Ablation}\label{app:snpc-ablation}

Table~\ref{tab:snpc-ablation} details macro-F1 for all 15 block combinations under the GRU SNPC (\textsc{ss-only}, $n{=}30$). Block~A ($+0.154$) and Block~D ($+0.107$) provide massive positive main effects (Table~\ref{tab:snpc-main-effects}). Crucially, Block~B (LIWC) yields a \emph{negative} main effect ($-0.041$), inverting its batch-setting behavior ($+0.044$). In batch classifiers, LIWC aggregation smooths out post-level variation. In sequential models, presenting unaggregated linguistic shifts step-by-step introduces severe noise, forcing the recurrent state to waste capacity discounting it.

\begin{table}[t]
\centering
\small
\setlength{\tabcolsep}{4pt}
\renewcommand{\arraystretch}{1.10}
\caption{SNPC main effects by feature block (GRU, \textsc{ss-only}, $n{=}30$). Main effect = mean macro-F1 with block present $-$ mean macro-F1 with block absent, marginalized over all other block combinations.}
\label{tab:snpc-main-effects}
\begin{tabular}{@{}lccc@{}}
\toprule
\textbf{Block} & \textbf{With block} & \textbf{Without block} & \textbf{Main effect} \\
\midrule
A (rates + counts, 20f)        & .646 & .493 & $+.154$ \\
D (excerpt positions, 10f)     & .625 & .518 & $+.107$ \\
C (SS prop.\ + words, 2f)      & .574 & .576 & $-.002$ \\
B (LIWC, 8f)                   & .555 & .597 & $-.041$ \\
\bottomrule
\end{tabular}
\end{table}

\begin{table}[t]
\centering
\small
\renewcommand{\arraystretch}{1.10}
\caption{Macro-F1 (mean $\pm$ SD across 5 stratified folds) for all 15 non-empty feature-block combinations at $n{=}30$ under the GRU SNPC, \textsc{ss-only} filtering, $h{=}128$. Best combination in \textbf{bold}.}
\label{tab:snpc-ablation}
\begin{tabular}{ccccccc}
\toprule
\textbf{A} & \textbf{B} & \textbf{C} & \textbf{D} & \textbf{Features} & \textbf{Macro-F1} & \textbf{SD} \\
\midrule
\cmark & \xmark & \xmark & \xmark & 20 & 0.724 & 0.054 \\
\xmark & \cmark & \xmark & \xmark & 8 & 0.375 & 0.085 \\
\xmark & \xmark & \cmark & \xmark & 2 & 0.319 & 0.031 \\
\xmark & \xmark & \xmark & \cmark & 10 & 0.656 & 0.022 \\
\addlinespace
\cmark & \cmark & \xmark & \xmark & 28 & 0.702 & 0.053 \\
\cmark & \xmark & \cmark & \xmark & 22 & 0.720 & 0.036 \\
\cmark & \xmark & \xmark & \cmark & 30 & \textbf{0.737} & 0.051 \\
\xmark & \cmark & \cmark & \xmark & 10 & 0.414 & 0.026 \\
\xmark & \cmark & \xmark & \cmark & 18 & 0.650 & 0.013 \\
\xmark & \xmark & \cmark & \cmark & 12 & 0.651 & 0.028 \\
\addlinespace
\cmark & \cmark & \cmark & \xmark & 30 & 0.689 & 0.040 \\
\cmark & \cmark & \xmark & \cmark & 38 & 0.693 & 0.043 \\
\cmark & \xmark & \cmark & \cmark & 32 & 0.720 & 0.056 \\
\xmark & \cmark & \cmark & \cmark & 20 & 0.646 & 0.036 \\
\addlinespace
\cmark & \cmark & \cmark & \cmark & 40 & 0.703 & 0.031 \\
\bottomrule
\end{tabular}
\end{table}

\subsubsection{Architecture Comparison: GRU vs.\ LSTM}\label{app:snpc-arch}

Table~\ref{tab:snpc-arch} reports macro-F1 for GRU and LSTM under the best configuration per architecture (\textsc{ss-only} filtering, $h{=}128$). The two architectures are interchangeable on aggregate performance: the maximum absolute difference across truncation points is $0.023$ (at $n{=}10$, favoring LSTM), and the difference at $n{=}30$ is below $0.001$. Per-persona F1 differences at $n{=}30$ are also small (within $0.015$; see Appendix~\ref{app:per-persona}).

\begin{table}[t]
\centering
\small
\renewcommand{\arraystretch}{1.10}
\caption{Macro-F1 by architecture across truncation points $n$, for the best feature-block configuration per architecture (GRU AD, LSTM AC), \textsc{ss-only} filtering, $h{=}128$. $\Delta$ = GRU $-$ LSTM.}
\label{tab:snpc-arch}
\begin{tabular}{cccc}
\toprule
$n$ & \textbf{GRU} & \textbf{LSTM} & $\Delta$ \\
\midrule
1 & 0.443 & 0.430 & +0.013 \\
3 & 0.506 & 0.486 & +0.020 \\
5 & 0.536 & 0.543 & -0.007 \\
10 & 0.609 & 0.631 & -0.023 \\
20 & 0.701 & 0.715 & -0.014 \\
30 & 0.737 & 0.736 & +0.001 \\
\bottomrule
\end{tabular}
\end{table}

\subsubsection{Hidden-Size Sensitivity}\label{app:snpc-hidden}

Table~\ref{tab:snpc-hidden} reports macro-F1 across hidden sizes $h \in \{32, 64, 128\}$. Performance is largely insensitive beyond $h{=}32$ (max mean difference $= 0.023$). We selected $h{=}128$ for the headline configuration due to marginally tighter variance at $n \geq 20$.

\begin{table}[t]
\centering
\small
\setlength{\tabcolsep}{3.5pt}
\renewcommand{\arraystretch}{1.10}
\caption{Hidden-size sensitivity: macro-F1 mean across 5 stratified folds for the GRU best configuration (A+D, \textsc{ss-only}). Each cell shows mean over SD.}
\label{tab:snpc-hidden}
\begin{tabular}{@{}c@{\hskip 6pt}cccccc@{}}
\toprule
\textbf{$h$} & $n{=}1$ & $n{=}3$ & $n{=}5$ & $n{=}10$ & $n{=}20$ & $n{=}30$ \\
\midrule
32  & \shortstack{.283 \\ {\scriptsize $\pm$.068}} & \shortstack{.401 \\ {\scriptsize $\pm$.108}} & \shortstack{.524 \\ {\scriptsize $\pm$.076}} & \shortstack{.545 \\ {\scriptsize $\pm$.118}} & \shortstack{.676 \\ {\scriptsize $\pm$.027}} & \shortstack{.729 \\ {\scriptsize $\pm$.057}} \\
\addlinespace
64  & \shortstack{.253 \\ {\scriptsize $\pm$.043}} & \shortstack{.452 \\ {\scriptsize $\pm$.102}} & \shortstack{.510 \\ {\scriptsize $\pm$.069}} & \shortstack{.622 \\ {\scriptsize $\pm$.036}} & \shortstack{.706 \\ {\scriptsize $\pm$.025}} & \shortstack{.706 \\ {\scriptsize $\pm$.052}} \\
\addlinespace
128 & \shortstack{.279 \\ {\scriptsize $\pm$.050}} & \shortstack{.489 \\ {\scriptsize $\pm$.082}} & \shortstack{.521 \\ {\scriptsize $\pm$.078}} & \shortstack{.605 \\ {\scriptsize $\pm$.046}} & \shortstack{.687 \\ {\scriptsize $\pm$.017}} & \shortstack{.709 \\ {\scriptsize $\pm$.038}} \\
\bottomrule
\end{tabular}
\vspace{2pt}

{\footnotesize At $n{=}1$, one fold across $h{=}32$ and $h{=}128$ returned no valid predictions (degenerate single-step input); means and SDs at $n{=}1$ are computed over 4 folds.}
\end{table}

\subsubsection{Best-Configuration Recovery Curve}\label{app:snpc-recovery}

Table~\ref{tab:snpc-recovery} provides the tabular companion to Figure~\ref{fig:recovery_curve}, detailing the expanding \textsc{ss-only} cohort sizes as $n$ increases.


\begin{table}[t]
\centering
\small
\renewcommand{\arraystretch}{1.10}
\caption{Recovery curve for the SNPC best configuration (GRU, A+D, \textsc{ss-only}, $h{=}128$): mean $\pm$ SD macro-F1 across 5 folds, with cohort sizes.}
\label{tab:snpc-recovery}
\begin{tabular}{ccccc}
\toprule
$n$ & \textbf{Mean} & \textbf{SD} & $N$ & $N_\text{excluded}$ \\
\midrule
1 & 0.296 & 0.096 & 125 & 814 \\
3 & 0.506 & 0.072 & 335 & 839 \\
5 & 0.498 & 0.078 & 462 & 712 \\
10 & 0.574 & 0.085 & 696 & 478 \\
20 & 0.695 & 0.040 & 923 & 251 \\
30 & 0.737 & 0.051 & 1023 & 151 \\
\bottomrule
\end{tabular}
\end{table}

\subsection{Few-Shot LLM Prompting}\label{app:llm}

\subsubsection{Setup}

We evaluated \texttt{gpt-5.4-mini} and \texttt{gpt-5.4-nano} under 8-shot chain-of-thought prompting (two exemplars per persona, temperature $=0$). The system message defines the 10 indicators, specifies each persona via a five-field schema, and mandates a four-step CoT classification: (i) classify each indicator's frequency as \textsc{consistent}, \textsc{occasional}, or \textsc{absent}; (ii) assess co-occurrence and overall emotional intensity; (iii) match the resulting signature to one of the four persona prototypes; (iv) return a single-line JSON object with fields \texttt{persona} and \texttt{rationale}. Test posts appear chronologically with qualitative time gaps; absolute dates are hidden. The verbatim prompt, privacy-scrubbed exemplars, and inference code are released at \url{https://github.com/[redacted-for-review]}.

\begin{table}[t]
\centering
\small
\renewcommand{\arraystretch}{1.10}
\caption{LLM inference configuration.}
\label{tab:llm-config}
\begin{tabular}{@{}l p{0.62\columnwidth}@{}}
\toprule
\textbf{Component} & \textbf{Value} \\
\midrule
Models           & \texttt{gpt-5.4-mini-2026-03-17}, \texttt{gpt-5.4-nano-2026-03-17} \\
Prompt           & 8-shot CoT (2 exemplars $\times$ 4 personas) \\
Temperature      & 0.0 \\
Output           & JSON: \texttt{\{persona, rationale\}} \\
Input conditions & \textsc{ss-only}, \textsc{all-posts} \\
Truncation       & First $n \in \{1, 3, 5, 10, 20, 30\}$ posts \\
CV scheme        & 5-fold stratified \\
\bottomrule
\end{tabular}
\end{table}

\subsection{Per-Persona Results}\label{app:per-persona}

Table~\ref{tab:per-persona-all} consolidates per-persona F1 across all 16 methods evaluated in Phase~2. Notably, LLMs occasionally outperform structured methods at $n{=}1$ (e.g., gpt-5.4-nano P1 at $n{=}1$: $0.548$), but this zero-shot structural advantage erodes completely as history length $n$ grows.



\begin{sidewaystable*}[p]
\centering
\scriptsize
\setlength{\tabcolsep}{3.5pt}
\renewcommand{\arraystretch}{1.10}
\caption{Per-method, per-persona F1 across all 16 Phase~2 methods at six truncation points $n \in \{1, 3, 5, 10, 20, 30\}$. Methods grouped by family (baseline, Bayesian accumulator, batch, SNPC, few-shot LLM, oracles). Best non-oracle, non-majority result per column in \textbf{bold}. Mean across 5 stratified folds; values shown to 2 decimal places to fit the column budget. Block labels: A = indicator rates + counts; B = LIWC; C = SS proportion + word count; D = excerpt positions.}
\label{tab:per-persona-all}
\begin{tabular}{l*{30}{c}}
\toprule
& \multicolumn{6}{c}{\textbf{Macro}} & \multicolumn{6}{c}{\textbf{P0}} & \multicolumn{6}{c}{\textbf{P1}} & \multicolumn{6}{c}{\textbf{P2}} & \multicolumn{6}{c}{\textbf{P3}} \\
\cmidrule(lr){2-7} \cmidrule(lr){8-13} \cmidrule(lr){14-19} \cmidrule(lr){20-25} \cmidrule(lr){26-31}
\textbf{Method} & $1$ & $3$ & $5$ & $10$ & $20$ & $30$ & $1$ & $3$ & $5$ & $10$ & $20$ & $30$ & $1$ & $3$ & $5$ & $10$ & $20$ & $30$ & $1$ & $3$ & $5$ & $10$ & $20$ & $30$ & $1$ & $3$ & $5$ & $10$ & $20$ & $30$ \\
\midrule
Majority & .16 & .16 & .16 & .15 & .15 & .15 & .63 & .63 & .63 & .63 & .62 & .61 & .00 & .00 & .00 & .00 & .00 & .00 & .00 & .00 & .00 & .00 & .00 & .00 & .00 & .00 & .00 & .00 & .00 & .00 \\
\addlinespace
Bayes M1 (uniform) & .21 & .21 & .23 & .25 & .31 & .32 & .56 & .53 & .53 & .52 & .52 & .52 & .00 & .00 & .03 & .09 & .20 & .23 & .29 & .33 & .36 & .37 & .41 & .42 & .00 & .00 & .00 & .00 & .11 & .12 \\
Bayes M2 (SS-cond.) & .23 & .31 & .37 & .42 & .49 & .52 & .58 & .56 & .59 & .61 & .63 & .65 & .04 & .12 & .16 & .24 & .36 & .40 & .28 & .34 & .38 & .38 & .40 & .42 & .03 & .24 & .37 & .46 & .57 & .61 \\
Bayes M3 (SS-only) & .17 & .29 & .36 & .44 & .52 & .57 & .44 & .49 & .53 & .59 & .67 & .71 & .04 & .11 & .15 & .21 & .30 & .34 & \textbf{.34} & .35 & .35 & .37 & .40 & .44 & .03 & .34 & .49 & .60 & \textbf{.72} & \textbf{.77} \\
\addlinespace
Batch (all posts, ABD+SVM) & .27 & .30 & .32 & .32 & .40 & .48 & .57 & .53 & .51 & .46 & .47 & .60 & .23 & .26 & .32 & .31 & .41 & .49 & .16 & .26 & .28 & .36 & .52 & .56 & .13 & .13 & .14 & .17 & .21 & .25 \\
Batch (SS-cond., ABCD+SVM) & .17 & .30 & .34 & .40 & .47 & .49 & .16 & .49 & .55 & .55 & .60 & .61 & .24 & .30 & .39 & .43 & .50 & .53 & .21 & .27 & .24 & .37 & .48 & .57 & .09 & .15 & .20 & .23 & .30 & .26 \\
\addlinespace
SNPC GRU $\cdot$ AD (SS-only) & .30 & \textbf{.51} & .50 & .57 & .69 & \textbf{.74} & .56 & \textbf{.66} & \textbf{.66} & .71 & .80 & \textbf{.82} & .34 & \textbf{.56} & .53 & .64 & .72 & .74 & .27 & \textbf{.40} & \textbf{.39} & .42 & \textbf{.60} & .64 & .00 & .42 & .41 & .54 & .66 & .75 \\
SNPC LSTM $\cdot$ AC (SS-only) & .31 & .49 & \textbf{.52} & \textbf{.62} & \textbf{.71} & .74 & .52 & .65 & .66 & \textbf{.73} & \textbf{.80} & .82 & .40 & .50 & \textbf{.55} & \textbf{.68} & \textbf{.73} & \textbf{.74} & .32 & .37 & .39 & \textbf{.46} & .59 & \textbf{.64} & .00 & \textbf{.43} & \textbf{.50} & \textbf{.63} & .72 & .74 \\
SNPC GRU $\cdot$ AC (all posts) & .26 & .33 & .38 & .47 & .58 & .65 & \textbf{.61} & .58 & .58 & .61 & .71 & .75 & .17 & .27 & .38 & .46 & .60 & .62 & .12 & .23 & .32 & .33 & .47 & .55 & \textbf{.14} & .23 & .26 & .48 & .53 & .68 \\
SNPC LSTM $\cdot$ AC (all posts) & .26 & .35 & .34 & .43 & .53 & .62 & .61 & .59 & .57 & .59 & .66 & .72 & .22 & .32 & .28 & .42 & .60 & .60 & .11 & .25 & .30 & .35 & .47 & .55 & .11 & .23 & .20 & .37 & .40 & .60 \\
\addlinespace
gpt-5.4-nano (SS-only) & .27 & .35 & .36 & .34 & .35 & .36 & .33 & .39 & .40 & .40 & .41 & .43 & \textbf{.55} & .47 & .47 & .44 & .45 & .43 & .21 & .25 & .28 & .23 & .26 & .27 & .00 & .28 & .29 & .29 & .27 & .30 \\
gpt-5.4-mini (SS-only) & \textbf{.33} & .34 & .33 & .34 & .35 & .35 & .38 & .36 & .37 & .38 & .40 & .41 & .52 & .48 & .46 & .43 & .47 & .44 & .28 & .18 & .21 & .23 & .23 & .21 & .13 & .36 & .29 & .33 & .30 & .32 \\
gpt-5.4-nano (all posts) & .26 & .28 & .29 & .27 & .29 & .30 & .54 & .48 & .47 & .38 & .35 & .31 & .22 & .24 & .24 & .25 & .27 & .30 & .20 & .28 & .32 & .33 & .37 & .39 & .10 & .12 & .14 & .13 & .17 & .18 \\
gpt-5.4-mini (all posts) & .21 & .19 & .19 & .20 & .22 & .24 & .36 & .22 & .21 & .18 & .15 & .14 & .22 & .23 & .20 & .22 & .27 & .31 & .15 & .21 & .25 & .27 & .30 & .33 & .12 & .11 & .12 & .13 & .15 & .17 \\
\addlinespace
Oracle (post-aggregated) & --- & --- & --- & --- & --- & .62 & --- & --- & --- & --- & --- & .77 & --- & --- & --- & --- & --- & .65 & --- & --- & --- & --- & --- & .73 & --- & --- & --- & --- & --- & .34 \\
Oracle (SS-conditional) & --- & --- & --- & --- & --- & .86 & --- & --- & --- & --- & --- & .88 & --- & --- & --- & --- & --- & .90 & --- & --- & --- & --- & --- & .83 & --- & --- & --- & --- & --- & .83 \\
\bottomrule
\end{tabular}
\end{sidewaystable*}

\subsection{Cross-Method Robustness}\label{app:robustness}

\paragraph{Fold-level variance.}
Twenty-four folds were flagged as outliers ($>1.5$ SD from the mean) across all methods and truncation points: accumulator M3 accounted for 8, the batch classifier for 7, accumulator M1 for 3, M2 for 2, and oracle classifiers for 4. Variance is driven primarily by P3 representation in the test fold ($N{=}58$; ${\sim}11{-}12$ users per fold): when P3 members are disproportionately held out, the training data lacks sufficient representation to recover the cluster. M3 and the batch classifier are most sensitive because both rely on SS-post features where P3's extreme but sparse profile is most vulnerable to sampling fluctuation.

\section{Phase 3 Supplementary Details}\label{app:phase3}

\subsection{LLM Generation Setup}\label{app:phase3-generation}

We generate 144 cells (24 stimuli $\times$ 3 models $\times$ 2 conditions) with one API call per cell, $K{=}1$ samples, and temperature $0.7$, aligning with the reference class for empathic response generation in this domain \citep{badawiWhenCanWe2026, wangchatthero2025}. The exact API snapshots are \texttt{gpt-5.4-mini-2026-03-17} \cite{openaiGPT54_2026}, \texttt{llama4-scout-instruct-basic}, and \texttt{gemini-3-flash-preview} (Google Vertex AI). Use of these APIs is consistent with each provider's research-use terms; outputs are seen only by expert raters. The \textsc{matched} system prompt is constructed from the per-persona response rubric (\S\ref{sec:phase3}) and includes three persona-matched few-shot examples; the \textsc{neutral} system prompt is a generic supportive instruction with no examples. Both arms produce a structured \texttt{<thinking>/<response>} output parsed deterministically; only \texttt{<response>} content is shown to raters. We leave \texttt{max\_tokens} at provider default and do not pin reasoning-channel behavior; each provider uses its default chat-completions mode. All prompts and generation code are released at \url{https://github.com/[redacted-for-review]} under the MIT license for research use only; the classifiers, rubrics, and persona-conditioning prompts are not intended for clinical decision-making or user-facing deployment..

\subsection{Evaluation Items}\label{app:phase3-items}
We recruited 8 clinicians and researchers with expertise in clinical psychology and public health via professional networks. Raters were not financially compensated; participation was solicited as expert consultation consistent with norms for clinical-evaluation studies in this area. The survey takes approximately 60 minutes to complete.

Table~\ref{tab:phase3-items} lists all items in the rater survey distributed via Qualtrics. Each response is rated on the 4 generic and 4 cross-cutting Likert items, plus the 2 Likert items specific to the matched persona of the stimulus (10 Likert items per response). Two additional items code descriptive features of the response (advice category and unsignaled crisis-line referral). A single pairwise-preference item is collected once per stimulus--model pair after both responses have been rated independently.

\begin{table*}[t]
\centering
\small
\setlength{\tabcolsep}{4pt}
\renewcommand{\arraystretch}{1.15}
\caption{Phase~3 evaluation instrument. Likert items use a 5-point scale (1~=~Strongly Disagree, 5~=~Strongly Agree). Persona-specific items are presented only when the stimulus's persona matches. Descriptive-coding items (ADV, CRR) capture response-level features for content auditing.}
\label{tab:phase3-items}
\begin{tabular}{@{}lp{0.16\textwidth}p{0.48\textwidth}p{0.22\textwidth}@{}}
\toprule
\textbf{ID} & \textbf{Name} & \textbf{Item text} & \textbf{Source} \\
\midrule
\multicolumn{4}{@{}l}{\textit{Generic support-quality items (Likert; all stimuli)}} \\
\midrule
G1 & Encouraging Elaboration & The response invited the user to share more through questions, openings, or invitations to continue. & \citet{kumarWhenLargeLanguage2026}; \citet{sharma-etal-2020-computational} \\
G2 & Validating Emotions & The response acknowledged the emotions the user expressed without minimizing or amplifying them. & \citet{kumarWhenLargeLanguage2026} \\
G3 & Stayed Focused on User & The response stayed focused on the user rather than shifting to the responder's own experiences or perspective. & \citet{kumarWhenLargeLanguage2026} \\
G4 & Engaged with Concerns & The response engaged fully with the user's concerns rather than dismissing or minimizing them. & \citet{kumarWhenLargeLanguage2026} \\
\midrule
\multicolumn{4}{@{}l}{\textit{Cross-cutting stigma-sensitive items (Likert; all stimuli)}} \\
\midrule
S1 & Voice-Preserving Language & The response stayed within the user's vocabulary, avoiding clinical or diagnostic terms the user did not introduce, and avoiding stigmatizing terms. Echoing the user's own self-description is not penalized. & \citet{kosylukMentalDistressLabel2024}; \citet{NIDA_WordsMatter}; \citet{linkModifiedLabelingTheory1989} \\
RM & Register Matching & The response matched the user's emotional register without importing intensity, urgency, or alarm beyond what the user expressed. & \citet{wang2025evaluatingllmpoweredchatbotcognitive}; \citet{} \\
SP & Response Specificity & The response engaged with the specific content of the user's message rather than relying on generic phrases that could apply to any self-stigma message. & \citet{iftikharHowLLMCounselors2025}; \citet{song_typing_2025} \\
NPE & Non-Performative Empathy & The response avoided generic empathic phrases (``thank you for sharing,'' ``so brave'') and formulaic affirmations, conveying engagement through specific attention rather than performance. & \citet{iftikharHowLLMCounselors2025}; \citet{arnaiz2025between} \\
\midrule
\multicolumn{4}{@{}l}{\textit{Persona-specific items (Likert; matched persona only)}} \\
\midrule
P0-1 & Holds Tension Without Resolving & The response held tensions in the user's message without steering toward a particular conclusion about their substance use. & \citet{miller2012motivational} \\
P0-2 & Honors Autonomy & The response treated decisions about substance use as the user's to make rather than advocating a particular path. & \citet{miller2012motivational}; \citet{hawkHarmReductionPrinciples2017a} \\
P1-1 & Engages Without Contradicting & The response engaged with the user's self-statements without arguing against or directly contradicting them. & \citet{linkModifiedLabelingTheory1989}; \citet{luomaSlowSteadyWins2012} \\
P1-2 & Acknowledges Without Probing & The response acknowledged what the user disclosed without probing for unstated content. & \citet{Quinn2009LivingWA}; \citet{song_typing_2025} \\
P2-1 & Selective Engagement & The response engaged substantively with one or two specific elements rather than attempting comprehensive reflection of multiple elements. & \citet{iftikharHowLLMCounselors2025}; \citet{miller2012motivational} \\
P2-2 & Tolerates Unresolved Distress & The response sat with the user's situation as unresolved rather than pivoting to optimism, solutions, or closure. & \citet{wang2025evaluatingllmpoweredchatbotcognitive}; \citet{iftikharHowLLMCounselors2025} \\
P3-1 & Respects Self-Description & The response engaged with the user's self-description as written, without correcting, reframing, or person-firsting their identity-language. & \citet{linkModifiedLabelingTheory1989}; \citet{song_typing_2025}; \citet{ashfordSubstanceUseRecovery2018}  \\
P3-2 & Engages Stated Content Practically & The response engaged the practical content of the user's message rather than pivoting to emotional concern or change-talk the user did not signal. & \citet{iftikharHowLLMCounselors2025}; \citet{hawkHarmReductionPrinciples2017a} \\
\midrule
\multicolumn{4}{@{}l}{\textit{Descriptive coding (all stimuli)}} \\
\midrule
ADV & Advice Category & Did the response offer concrete advice or suggestions? Categories: None~$\vert$~Harm-reduction~$\vert$~Treatment referral~$\vert$~Behavior change~$\vert$~Practical-logistical~$\vert$~Other. & \citet{kumarWhenLargeLanguage2026}; \citet{yinAICanHelp2024} \\
CRR & Unsignaled Crisis-Line Referral & Did the response include a 988, SAMHSA, or other crisis-line referral when the user did not signal active crisis? (Yes / No) & \citet{arnaiz2025between}; \citet{SAMHSA2024} \\
\midrule
\multicolumn{4}{@{}l}{\textit{Pairwise preference (one per stimulus--model pair)}} \\
\midrule
PAIR-1 & Holistic Preference & Considering both responses, which response better engaged with this user's message overall? (A is better $\vert$ B is better $\vert$ Equivalent) & \citet{kumarWhenLargeLanguage2026} \\
\bottomrule
\end{tabular}
\end{table*}

\subsection{Analysis Details}\label{app:phase3-analysis}

\textbf{Model specifications.} Per-item Likert ratings were modeled with linear mixed-effects: \texttt{lmer(rating $\sim$ condition + (1|rater) + (1|stimulus))}, with random intercepts for rater and stimulus. The forced-choice item was modeled with a corresponding ordinal cumulative link mixed model (CLMM). Descriptive codes (advice category, unsignaled crisis-line referrals) were analyzed with Fisher's exact tests and binomial GLMMs with the same random-effects structure. LLM and persona were entered as secondary interaction terms.

\textbf{Pre-specified directions.} Table~\ref{tab:phase3-direction} gives the direction assigned to each of the 19 confirmatory tests. Universal items were predicted matched~$>$~neutral; G1 varied by persona because elaboration-probing was contraindicated for P1's shame-concealment profile and not pre-specified for P2 or P3; persona-specific items were predicted matched~$>$~neutral on their matching persona. Directional tests were one-sided in the predicted direction; the two G1 cells with no pre-specified direction were two-sided. All 19 raw $p$-values entered a single Benjamini--Hochberg FDR family at $q = .05$.

\begin{table}[t]
\centering
\small
\setlength{\tabcolsep}{6pt}
\renewcommand{\arraystretch}{1.1}
\begin{tabular}{@{}lcccc@{}}
\toprule
\textbf{Item(s)} & \textbf{P0} & \textbf{P1} & \textbf{P2} & \textbf{P3} \\
\midrule
G2, G3, G4, S1, RM, SP, NPE & $+$ & $+$ & $+$ & $+$ \\
G1 & $+$ & $-$ & $0$ & $0$ \\
P0-1, P0-2 & $+$ & n/a & n/a & n/a \\
P1-1, P1-2 & n/a & $+$ & n/a & n/a \\
P2-1, P2-2 & n/a & n/a & $+$ & n/a \\
P3-1, P3-2 & n/a & n/a & n/a & $+$ \\
\bottomrule
\end{tabular}
\caption{Pre-specified direction matrix for the 19 confirmatory tests. $+$ = matched~$>$~neutral predicted (one-sided); $-$ = matched~$<$~neutral predicted (one-sided); $0$ = no pre-specified direction (two-sided); n/a = item not rated on this persona's stimuli.}
\label{tab:phase3-direction}
\end{table}

\noindent\textbf{Composite construction.} The per-response Likert composite is the arithmetic mean of the ten applicable items: four generic (G1, G2, G3, G4), four cross-cutting (S1, RM, SP, NPE), and the two persona-specific items matching the stimulus's persona. G1 was reverse-coded ($6 - v$) on P1 stimuli to align with its pre-specified $-$ direction; G1 entered raw on P0, P2, and P3. The P3-only CRR analysis required a penalized GLM with weakly-informative priors \citep{Gelman_2008} due to complete separation (i.e., all 36 matched responses received CRR = No) to produce a finite estimate.


\subsection{Per-Item Confirmatory Effects}\label{app:phase3-peritem}
 
Table~\ref{tab:phase3-peritem-full} reports the full 19-test family referenced in the body. Tests are grouped by item type. Bold rows pass BH-FDR at $q = .05$ in the pre-specified direction (Table~\ref{tab:phase3-direction}).
 
\begin{table}[t]
\centering
\small
\setlength{\tabcolsep}{4pt}
\renewcommand{\arraystretch}{1.05}
\begin{tabular}{@{}l c r@{\hspace{2pt}}l c@{}}
\toprule
\textbf{Item} & \textbf{Pred.} & \multicolumn{2}{c}{\textbf{$\beta$ (95\% CI)}} & \textbf{$q$} \\
\midrule
\multicolumn{5}{@{}l}{\textit{Universal items}} \\
G2  & $+$ & $-0.67$ & $[-0.88, -0.46]$ & $\geq.99$ \\
G3  & $+$ & $-0.24$ & $[-0.38, -0.10]$ & $\geq.99$ \\
G4  & $+$ & $-0.59$ & $[-0.77, -0.41]$ & $\geq.99$ \\
S1  & $+$ & $+0.05$ & $[-0.12, +0.21]$ & $\phantom{<}.488$ \\
RM  & $+$ & $+0.06$ & $[-0.14, +0.25]$ & $\phantom{<}.488$ \\
SP  & $+$ & $-0.18$ & $[-0.36, -0.00]$ & $\geq.99$ \\
\textbf{NPE} & $+$ & $\mathbf{+0.53}$ & $\mathbf{[+0.30, +0.76]}$ & $\mathbf{<.001}$ \\
\addlinespace
\multicolumn{5}{@{}l}{\textit{G1 by persona}} \\
G1 $\times$ P0 & $+$ & $-1.17$ & $[-1.71, -0.63]$ & $\geq.99$ \\
\textbf{G1 $\times$ P1} & $-$ & $\mathbf{-1.42}$ & $\mathbf{[-1.92, -0.91]}$ & $\mathbf{<.001}$ \\
\textbf{G1 $\times$ P2} & $0$ & $\mathbf{-1.09}$ & $\mathbf{[-1.56, -0.62]}$ & $\mathbf{<.001}$ \\
\textbf{G1 $\times$ P3} & $0$ & $\mathbf{-1.69}$ & $\mathbf{[-2.23, -1.16]}$ & $\mathbf{<.001}$ \\
\addlinespace
\multicolumn{5}{@{}l}{\textit{Persona-specific items}} \\
\textbf{P0-1} & $+$ & $\mathbf{+0.56}$ & $\mathbf{[+0.15, +0.96]}$ & $\mathbf{\phantom{<}.012}$ \\
P0-2 & $+$ & $+0.25$ & $[-0.21, +0.71]$ & $\phantom{<}.307$ \\
P1-1 & $+$ & $-0.19$ & $[-0.54, +0.16]$ & $\geq.99$ \\
P1-2 & $+$ & $-0.03$ & $[-0.28, +0.23]$ & $\phantom{<}.853$ \\
\textbf{P2-1} & $+$ & $\mathbf{+0.79}$ & $\mathbf{[+0.34, +1.25]}$ & $\mathbf{\phantom{<}.002}$ \\
\textbf{P2-2} & $+$ & $\mathbf{+1.18}$ & $\mathbf{[+0.66, +1.69]}$ & $\mathbf{<.001}$ \\
P3-1 & $+$ & $+0.33$ & $[-0.18, +0.85]$ & $\phantom{<}.248$ \\
P3-2 & $+$ & $-0.03$ & $[-0.42, +0.37]$ & $\phantom{<}.853$ \\
\bottomrule
\end{tabular}
\caption{Per-item $\beta$ (matched $-$ neutral) from \texttt{lmer(rating $\sim$ condition + (1$\mid$rater) + (1$\mid$stimulus))}. \textbf{Pred.}: pre-specified direction; \textbf{$q$}: BH-FDR across all 19 tests. Bold rows pass FDR in the pre-specified direction.}
\label{tab:phase3-peritem-full}
\end{table}

\subsection{Inter-Rater Reliability}\label{app:phase3-irr}
 
Each response was rated by two of eight experts under a partially cross-reference assignment. Table~\ref{tab:phase3-irr} reports two reliability metrics per item: Gwet's AC2 (ordinal for Likert items, AC1 unweighted for the nominal ADV and binary CRR codes), which is robust to skewed marginal distributions \citep{gwet2014handbook, feinstein1990high}, and observed agreement within-one-point on the Likert scale (exact agreement for ADV and CRR). CRR achieves strong reliability across both metrics; ADV is moderate; Likert items are mixed.
 
\begin{table}[t]
\centering
\small
\setlength{\tabcolsep}{4pt}
\renewcommand{\arraystretch}{1.05}
\begin{tabular}{@{}l r r@{}}
\toprule
\textbf{Item} & \textbf{AC} & \textbf{w-1} \\
\midrule
\multicolumn{3}{@{}l}{\textit{Generic + cross-cutting Likert}} \\
G1  & $0.25$ & $0.68$ \\
\textbf{G2}  & $\mathbf{0.64}$ & $\mathbf{0.77}$ \\
G3  & $0.47$ & $0.83$ \\
\textbf{G4}  & $\mathbf{0.70}$ & $\mathbf{0.81}$ \\
\textbf{S1}  & $\mathbf{0.69}$ & $\mathbf{0.79}$ \\
RM  & $0.33$ & $0.73$ \\
SP  & $0.17$ & $0.74$ \\
NPE & $0.12$ & $0.69$ \\
\addlinespace
\multicolumn{3}{@{}l}{\textit{Persona-specific Likert}} \\
P0-1 & $0.00$ & $0.61$ \\
P0-2 & $\phantom{-}0.32$ & $0.67$ \\
P1-1 & $\phantom{-}0.51$ & $0.83$ \\
\textbf{P1-2} & $\phantom{-}\mathbf{0.72}$ & $\mathbf{0.83}$ \\
P2-1 & $\phantom{-}0.06$ & $0.86$ \\
P2-2 & $\phantom{-}0.31$ & $0.61$ \\
P3-1 & $\phantom{-}0.00$ & $0.67$ \\
P3-2 & $-0.21$ & $0.61$ \\
\addlinespace
\multicolumn{3}{@{}l}{\textit{Descriptive + pairwise}} \\
ADV    & $0.42$ & $0.50$ \\
\textbf{CRR} & $\mathbf{0.91}$ & $\mathbf{0.92}$ \\
\textbf{PAIR-1} & $\mathbf{0.73}$ & $\mathbf{0.78}$ \\
\bottomrule
\end{tabular}
\caption{Inter-rater reliability. \textbf{AC}: Gwet's AC2 (ordinal) for Likert and PAIR-1; Gwet's AC1 (unweighted) for ADV and CRR. \textbf{w-1}: proportion of rater pairs within one scale point (Likert and PAIR-1) or exact agreement (ADV and CRR). \textbf{Bold:} achieves substantial reliability across both metrics.}
\label{tab:phase3-irr}
\end{table}

\end{document}